\documentclass[runningheads]{llncs}

\usepackage{eccv}

\usepackage{eccvabbrv}

\usepackage{graphicx}
\usepackage{booktabs}

\usepackage{graphicx}
\usepackage{subcaption}
\usepackage[breaklinks,colorlinks,allcolors=cvprblue]{hyperref}
\newcommand{\cmark}{\checkmark}
\newcommand{\xmark}{\texttimes}
\usepackage[table]{xcolor}
\usepackage{amsmath}
\usepackage{amssymb}
\usepackage[accsupp]{axessibility}  
\usepackage{xcolor}
\definecolor{cvprblue}{rgb}{0.21,0.49,0.74}

\usepackage{hyperref}

\usepackage{orcidlink}

\begin{document}

\title{Moving Beyond More Views: Redundancy-Aware Ego–Exo Fusion for Proficiency Estimation} 

\titlerunning{Redundancy-Aware Ego–Exo Fusion for Proficiency Estimation}

\author{Xu Dong\inst{1}\orcidlink{0009-0006-6842-8750} \and
Wanqing Li\inst{2}\orcidlink{0000-0002-4427-2687} \and
Anthony Adeyemi-Ejeye\inst{1}\orcidlink{0000-0002-8371-7829} \and
Andrew Gilbert\inst{1}\orcidlink{0000-0003-3898-0596}}

\authorrunning{X.Dong et al.}

\institute{University of Surrey, Guildford, United Kingdom\\ \and
University of Wollongong, Wollongong, Australia\\
\email{x.dong@surrey.ac.uk, a.gilbert@surrey.ac.uk} }

\maketitle

\begin{abstract}
  \noindent EgoExo proficiency estimation aims to assess action quality by integrating fine-grained motion cues from egocentric (1st-person) views with spatial context from multiple exocentric (3rd-person) views. Simply adding more exocentric views degrades EgoExo performance, as redundant or noisy perspectives dilute useful motion cues. Our analysis identifies two key causes:
\textbf{(1) Multiview redundancy} — From the data perspective, certain views provide limited or noisy information, diluting discriminative cues;
\textbf{(2) Overfitting} — From the feature perspective, conventional fusion increases representational complexity, causing the model to memorise view-specific patterns rather than learn generalisable representations. To address these issues, we propose two complementary modules.
\emph{AdaMVS} adaptively identifies and fuses the most informative view tokens under weak supervision from the data perspective, while \emph{VIB-GB} combines Gradient Blending and Variational Information Bottleneck regularisation from the feature perspective to compress redundant signals and suppress overfitting during training.
Experiments on EgoExo-4D and EgoExo-Fitness demonstrate that our method learns both \textbf{which view to look at} and \textbf{how to fuse them}, achieving new state-of-the-art results. Our source code is available at \url{https://github.com/dx199771/AdaMVS}
\keywords{Action Quality Assessment \and Egocentric and Exocentric Vision \and Multiview Learning \and Redundancy-Aware Fusion \and View Selection}
\end{abstract}

\begin{figure}[t]
    \centering
    \begin{subfigure}{0.54\linewidth}
        \centering
        \includegraphics[width=\linewidth]{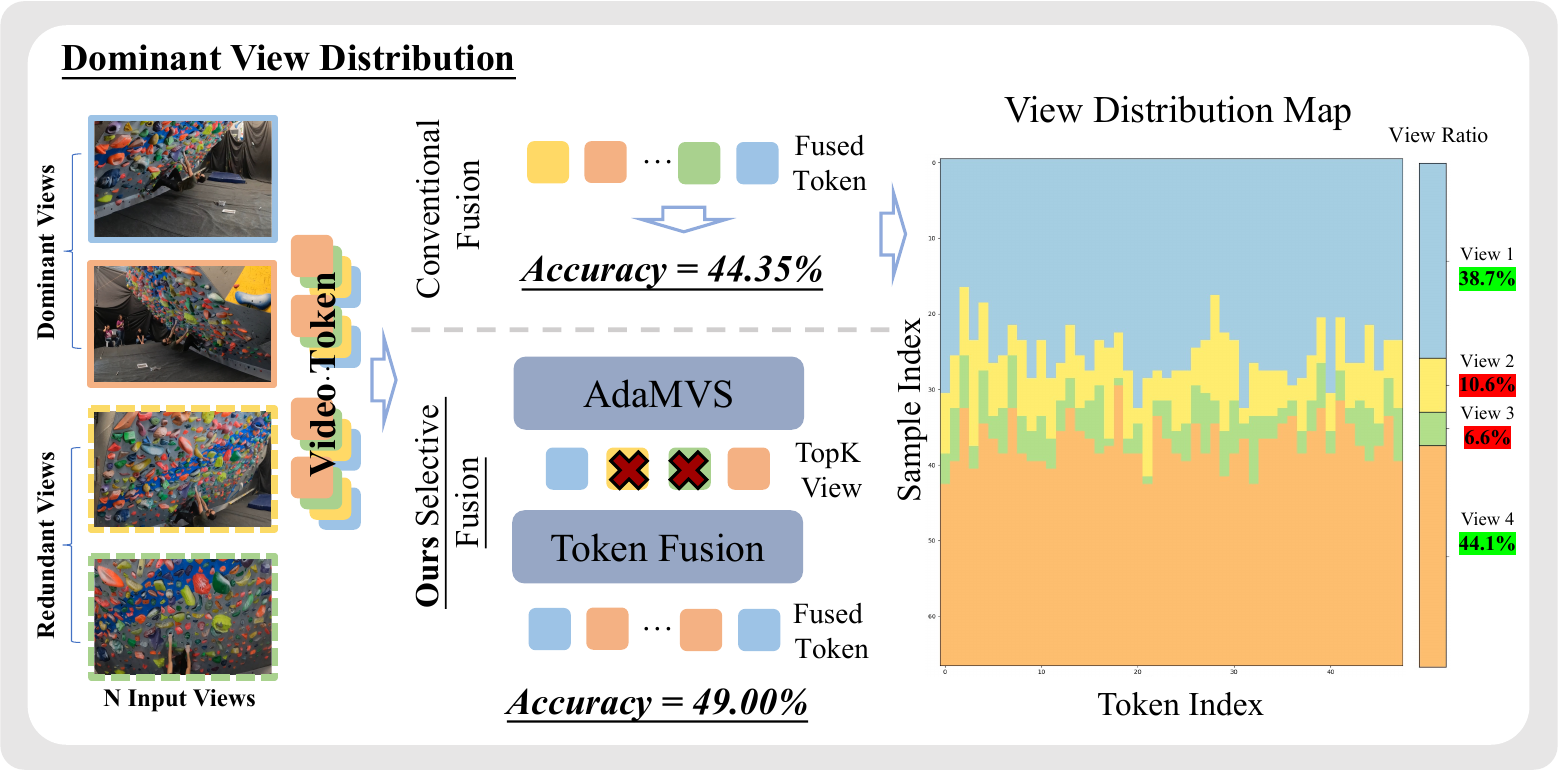}
        \caption{Dominant View Distribution.}
        \label{fig:teaser:a}
    \end{subfigure}
    \hfill
    \begin{subfigure}{0.45\linewidth}
        \centering
        \includegraphics[width=\linewidth]{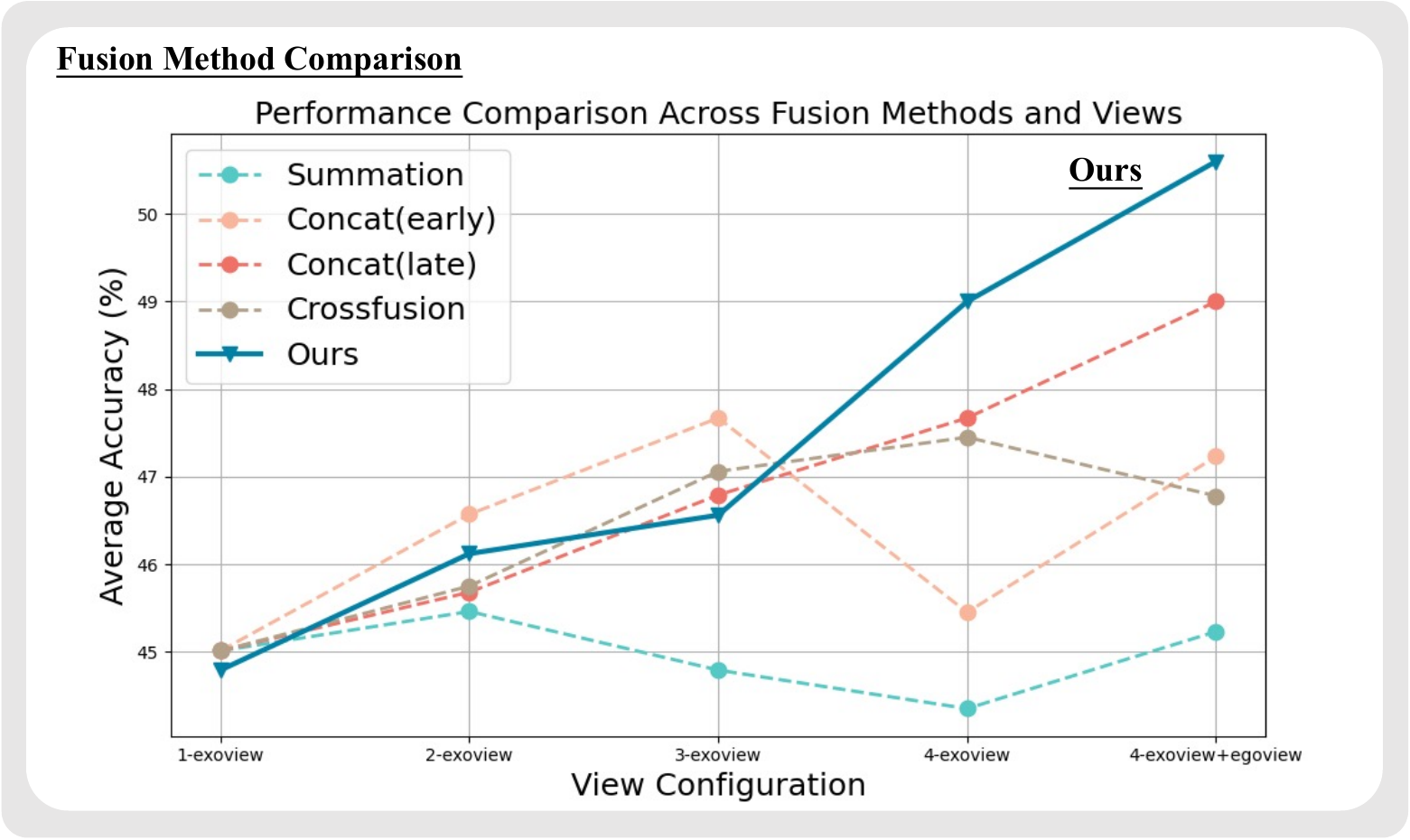}
        \caption{Fusion Method Comparison.}
        \label{fig:teaser:b}
    \end{subfigure}
    \caption{\textbf{Multi-Exo-View Fusion Challenges.} \protect\subref{fig:teaser:a} demonstrates that redundant views  (e.g., Views 2 and 3 account for just \textbf{17.2\%} of the total attention weight) contribute only marginal information; \protect\subref{fig:teaser:b} shows that simply adding more views can degrade overall performance. In contrast, our method adaptively identifies and fuses the most informative views to enhance robustness and accuracy.}
    \label{fig:teaser}
\end{figure}

\section{Introduction}
\label{sec:intro}

Proficiency estimation aims to accurately and efficiently evaluate human skill levels, encompassing tasks such as assessing athletes’ training effectiveness, monitoring everyday functional abilities (e.g., cooking or repairing), and evaluating professional training in domains like music or medicine with contextual understanding.
Unlike conventional \emph{Action Quality Assessment} (AQA) methods~\cite{pirsiavash2014assessing,yu2021groupawarecontrastiveregressionaction,doughty2019prosconsrankawaretemporal,Xu_2024_CVPR_fineparser,zhang2024logolongformvideodataset}, that rely solely on third-person footage, EgoExo learning integrates egocentric motion cues with multi-view spatial context~\cite{grauman2024egoexo4dunderstandingskilledhuman,li2024egoexofitnessegocentricexocentricfullbody}, enabling fine-grained proficiency and richer contextual understanding of human skill. Specifically, egocentric videos capture subtle motion details and intention cues (e.g., hand–object interactions). At the same time, exocentric views provide complementary information on full-body dynamics and spatial context (e.g., body posture in sports).
This promising paradigm demonstrates the potential of leveraging multi-view information for proficiency estimation, while also revealing the challenges inherent in effectively utilising such diverse perspectives.

Therefore, efficiently exploiting multi-view information becomes a key problem. Ego–Exo fusion differs from standard multi-camera setups: viewpoints are heterogeneous and unaligned, making static fusion brittle. Although multiview learning has been widely studied in other domains, these methods are not robust for \emph{EgoExo learning}. They often adopt static fusion strategies~\cite{hamdi2021mvtnmultiviewtransformationnetwork, su2015multiviewconvolutionalneuralnetworks}, which fail to suppress noisy or uninformative views and thus cannot effectively mitigate information redundancy. In addition, general multimodal fusion approaches~\cite{wang2020deepmultimodalfusionchannel, jia2024geminifusionefficientpixelwisemultimodal} struggle to handle the inherent feature divergence and cross-view overfitting in Ego–Exo data. In our experiments, conventional multiview fusion fails to yield consistent improvements with more views (adding one extra view even resulted in a 1–2\% accuracy drop), leading to degraded overall performance (\cref{fig:teaser}). We analyse and summarise the primary causes of this degradation from two complementary perspectives: \textbf{data} and \textbf{feature}. \textbf{Redundancy:} More views are not always better; additional exocentric \emph{data} viewpoints often introduce occlusions or duplicate content, which hurt generalisation (\cref{fig:teaser:b}). Conventional fusion methods \cite{Ngiam2011MultimodalDL,tsai2019multimodal} fail to adaptively weight or suppress uninformative views (\cref{fig:teaser:a}), giving redundant ones excessive attention and introducing noise that weakens useful features. \textbf{Overfitting:} The enlarged \emph{feature} space amplifies branch imbalance, causing the network to memorise view-specific noise, especially when data are limited. This is particularly evident in Ego-Exo scenarios, where heterogeneous feature distributions and uneven convergence across views cause the model to memorise view-specific patterns rather than learn generalisable representations.

Empirically, multiview redundancy and overfitting tend to co-occur: redundancy in data is often accompanied by increased overfitting in the feature space, and excessive model adaptation may in turn amplify redundant signals. These factors jointly contribute to the observed performance degradation. To effectively address these two issues and enhance the robustness of multiview fusion, we design our framework from two complementary perspectives. At the data level, we propose \textbf{AdaMVS} (\emph{Adaptive Multiview Selector}), which employs an adaptive scoring mechanism to dynamically identify and select the most informative Top-$K$ exocentric views and fuse by token exo-fusion. These are then integrated with the egocentric stream to effectively reduce redundancy and computational cost.
At the feature level, we introduce a complementary \textbf{VIB-GB} module that combines the \emph{Variational Information Bottleneck (VIB)} and \emph{Gradient Blending (GB)}. By compressing non-essential signals and balancing gradient flow, it mitigates feature overfitting, preventing the model from memorising view-specific patterns and enabling the learning of compact, robust, and generalisable cross-view representations.

Together, AdaMVS and VIB-GB form a unified framework that jointly learns \emph{what to fuse} and \emph{how to regularise fusion}.
\noindent \textbf{Contributions.}
\begin{itemize}
\item We identify and analyse two critical issues in EgoExo fusion: \textbf{(1)} multiview information redundancy and \textbf{(2)} strong susceptibility to overfitting.
\item We introduce an adaptive token-level multiview selector, \textbf{AdaMVS}, and couple it with an OGR-driven gradient reweighting module, \textbf{VIB-GB}, forming a unified framework that adaptively reduces redundancy and explicitly controls overfitting.
\item Experiments on \textbf{EgoExo-4D} and \textbf{EgoExo-Fitness} show state-of-the-art proficiency estimation with improved robustness and efficiency.
\end{itemize}

\section{Related Work}
\paragraph{Egocentric and Exocentric Understanding}

\noindent Egocentric and exocentric videos offer complementary perspectives critical for skill understanding: first-person views capture fine-grained hand–object interactions and attention cues, while third-person views provide holistic body motion and scene context. Most prior work focuses on one perspective—either egocentric understanding~\cite{pan2023egovitpyramidvideotransformer,wang2023egoonlyegocentricactiondetection,sudhakaran2019lstalongshorttermattention,grauman2022ego4dworld3000hours,damen2018scalingegocentricvisionepickitchens,zhang2022finegrainedegocentrichandobjectsegmentation,kevin2022egovlp} or exocentric video analysis~\cite{kay2017kineticshumanactionvideo,monfort2019momentstimedatasetmillion}.
Joint learning across views remains challenging due to viewpoint gaps~\cite{Luo_2025_CVPR} and redundant cross-view information~\cite{majumder2025viewpointshowsbestlanguage}.

Recent Ego–Exo datasets~\cite{grauman2024egoexo4dunderstandingskilledhuman,li2024egoexofitnessegocentricexocentricfullbody,huang2025egoexolearndatasetbridgingasynchronous,elfeki2018personpersondatasetbaselines} enable multi-view tasks including proficiency estimation~\cite{grauman2024egoexo4dunderstandingskilledhuman,li2024egoexofitnessegocentricexocentricfullbody}, action anticipation~\cite{huang2025egoexolearndatasetbridgingasynchronous}, person localisation~\cite{9607539}, and 3D pose estimation~\cite{wang2022estimatingegocentric3dhuman,grauman2024egoexo4dunderstandingskilledhuman}.
Some approaches transfer knowledge from exocentric to egocentric views~\cite{lifting,li2021egoexotransferringvisualrepresentations}, while others seek view-invariant features~\cite{Luo_2025_CVPR,xue2023learningfinegrainedviewinvariantrepresentations,9220850,11098892} through cross-view alignment.
However, their scalability is constrained by limited paired data.
AE2 \cite{xue2023learningfinegrainedviewinvariantrepresentations} introduces self-supervised temporal alignment to learn fine-grained view-invariant actions, yet view invariance alone cannot address redundancy or noise from uninformative views.
LangView~\cite{majumder2025viewpointshowsbestlanguage} further explores viewpoint importance through weakly supervised language-guided selection.
While prior work seeks view invariance or uses coarse, language-guided selection, they fail to address the underlying noise and redundancy issues. Inspired by these limitations, our method performs token-level view selection to evaluate and fuse the most salient ego-exo information adaptively.

\begin{figure*}[t]
    \centering
    \includegraphics[width=1\linewidth]{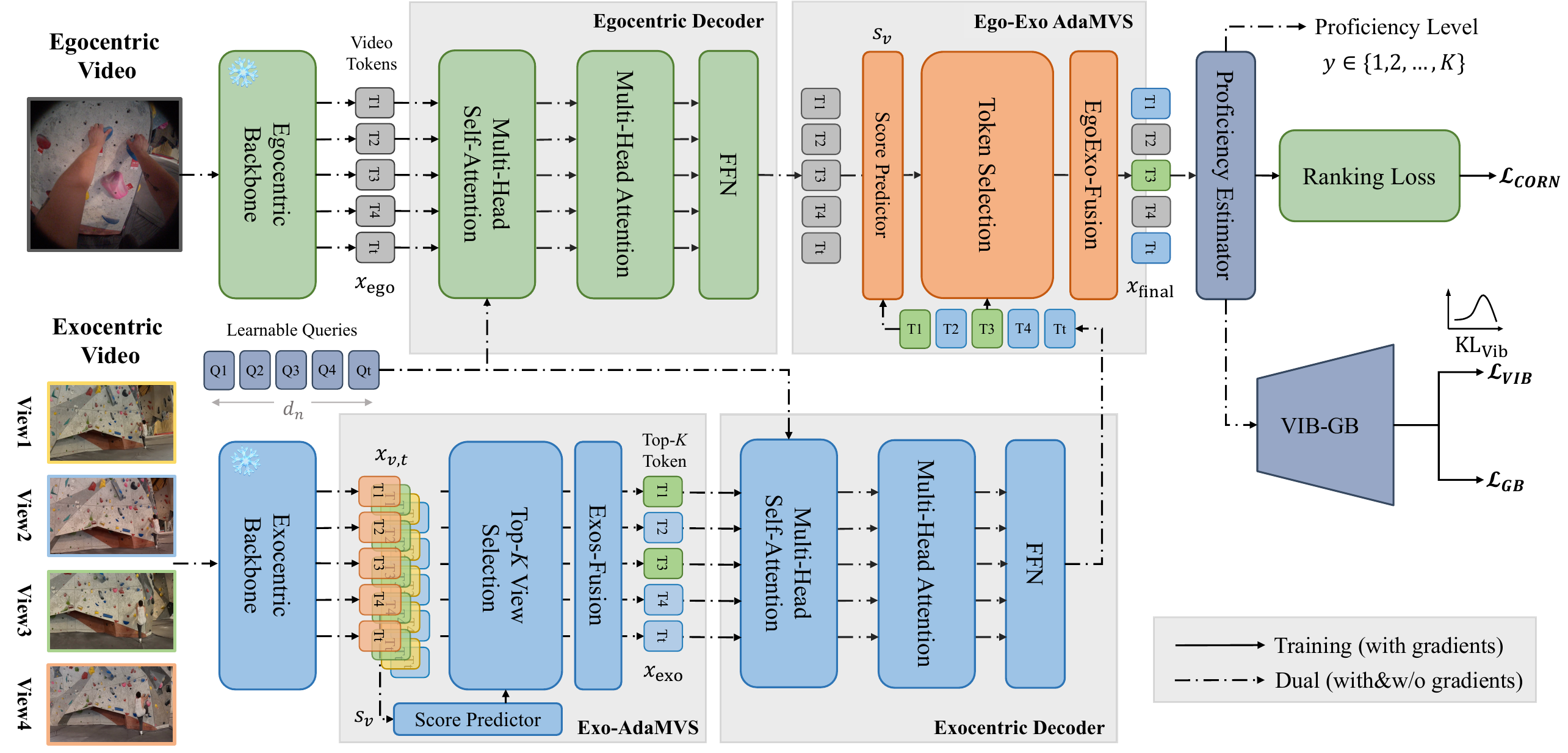}

    \caption{Overview of our proposed network.
    \textbf{(Top)} The egocentric branch extracts and decodes features through a dedicated pathway.
    \textbf{(Bottom)} The exocentric branch uses the Score Predictor and Exo-AdaMVS to compute scores, selecting key views and removing redundancy. The Ego–Exo AdaMVS fuses streams, while the VIB-GB module regularises the latent space to reduce overfitting and enhance generalisation. Solid arrows denote gradient flow, and dash-dotted arrows indicate mixed paths involving gradient and non-gradient operations used by VIB-GB.}

    \label{fig:network}
\end{figure*}
\paragraph{Proficiency Estimation}

\noindent Proficiency estimation is closely related to \emph{Action Quality Assessment} (AQA): AQA predicts continuous quality scores, while proficiency estimation classifies discrete skill levels.
Early AQA works relied on handcrafted features~\cite{pirsiavash2014assessing}, later replaced by CNN, GNN, and Transformer architectures for spatio-temporal reasoning~\cite{yu2021groupawarecontrastiveregressionaction,doughty2019prosconsrankawaretemporal,zhou2025uncertaintydrivenactionqualityassessment,dong2024interpretablelongtermactionquality,zhou2024cofinal,zhang2024logolongformvideodataset}.
Pose-based methods~\cite{Xu_2024_CVPR_fineparser,parmar2017learning} neglect contextual cues, whereas \emph{EgoExo4D}~\cite{grauman2024egoexo4dunderstandingskilledhuman} reframes the task by combining egocentric object interactions with exocentric spatial context.
Existing methods~\cite{grauman2024egoexo4dunderstandingskilledhuman,fan2020pyslowfast,rw2019timm} primarily rely on late fusion, overlooking temporal variations in view importance.
SkillFormer~\cite{bianchi2025skillformerunifiedmultiviewvideo} introduces efficient cross-view fusion and achieves strong results on EgoExo4D. However, effectively fusing these heterogeneous Ego-Exo streams for proficiency estimation remains uniquely challenging, as it requires not only managing multiview redundancy but also mitigating severe overfitting risks caused by heterogeneous inputs. Our approach addresses these issues through adaptive token-level fusion that emphasises discriminative views and regularised training that enhances generalisation.

\paragraph{Multi-view Learning}

Multi-view learning underpins domains like 3D object or pose~\cite{Chharia_2025_CVPR,hamdi2021mvtnmultiviewtransformationnetwork, hou2024learning}, anomaly detection~\cite{liu2025multiviewindustrialanomalydetection}, and action analysis~\cite{bianchi2025skillformerunifiedmultiviewvideo,huang2025egoexolearndatasetbridgingasynchronous,fan2022dribo, nguyen2024action}. The main challenges are feature alignment and robust generalisation. Existing pixel/token fusion methods~\cite{jia2024geminifusionefficientpixelwisemultimodal,wang2020deepmultimodalfusionchannel,wang2022multimodaltokenfusionvision, yan2022multiview} are often costly or fail to capture the dynamically varying informational value of each view. This is critical in Ego–Exo fusion, where differing perspectives exacerbate overfitting risks. Multiview systems are prone to overfitting because views generalise at varying rates~\cite{tejero2024complementary}. We quantify this imbalance using the Overfitting-to-Generalisation Ratio (OGR)~\cite{wang2020makestrainingmultimodalclassification}.
Our approach minimises OGR by combining adaptive selection with information-theoretic regularisation. This paradigm jointly addresses feature redundancy and overfitting, leading to generalisable Ego–Exo fusion.

In summary, despite progress in cross-view understanding and proficiency estimation, a unified solution is still needed to jointly address feature redundancy and overfitting in heterogeneous Ego–Exo data. Our AdaMVS framework, with the VIB-GB module, bridges this gap for robust Ego–Exo proficiency estimation.

\section{Methodology}

In this work, we address the task of proficiency estimation from multiview ego–exo videos, aiming to predict action proficiency labels (\cref{sec:1}). As illustrated in \cref{fig:network}, our framework first employs a frozen backbone to extract spatiotemporal features from each view. The multiple exocentric views are then processed by an Exocentric Adaptive Multiview Selector (Exo-AdaMVS) (\cref{par:2}), which selects the \text{Top-}\emph{K} most informative views to mitigate redundancy and capture representative temporal dynamics through an Exocentric Decoder. In parallel, the egocentric stream is encoded by an Egocentric Decoder to extract fine-grained motion cues. The ego and \text{Top-}\emph{K} exo tokens are fused through an Ego-Exo AdaMVS (\cref{par:2}), and the fused representation is further regularised by the VIB-GB mechanism (\cref{sec:3}), which integrates a Variational Information Bottleneck with Gradient Blending regularisation to alleviate overfitting. The optimisation strategy is detailed in \cref{sec:4}.

\subsection{Problem Formulation} 
\label{sec:1}
We formulate multiview Proficiency Estimation as an ordinal classification task using synchronised Ego-Exo videos. 
Given video sequences $V = \{v_{\text{ego}}, \allowbreak v_{\text{exo}}^{(1)}, \allowbreak v_{\text{exo}}^{(2)}, \dots, \allowbreak v_{\text{exo}}^{(n)}\}$,
where \( v_{\text{ego}} \) denotes the egocentric view and \( v_{\text{exo}}^{(i)} \) the \( i \)-th exocentric view, 
the goal is to predict a discrete proficiency label \( y \in \{1, 2, \dots, K\} \), with \( K \) indicating the number of skill levels. We learn a fusion function \( f: V \rightarrow y \) that integrates multiview information and maps the fused representation to a proficiency level. Unlike \emph{Action Quality Assessment (AQA)}, which performs regression on continuous scores, 
Proficiency Estimation focuses on discrete classification, aligning better with standard assessment protocols.

\subsection{AdaMVS}

Different exocentric views often contain overlapping or noisy cues, leading to uninformative information. To mitigate this redundancy, our framework employs an Adaptive Multiview Selector (AdaMVS), as shown in \cref{fig:network}, which selectively identifies the most informative views. AdaMVS consists of two branches: one for egocentric video and another for multiple exocentric videos, both built on a transformer-based architecture (see the supplementary material for details). The branches share an initial feature extraction stage where each video is uniformly sampled into $T$ clips and encoded by a frozen pretrained backbone to obtain feature representations. A shared set of learnable queries $Q$ acts as semantic anchors to enforce a common latent space, allowing each corresponding clip token to capture richer temporal and semantic dependencies. AdaMVS is inherently view-number agnostic due to the permutation-invariant Transformer and shared queries.

\paragraph{Exocentric Branch}\label{par:2}


Given multiple exocentric inputs, each view $v$ yields token-level clip features $x_{v,t}$ ($t$ for temporal index). Each token feature $x_{v,t}$ passes through a transformer-based score predictor $\phi(\cdot)$ producing importance weights aggregated per view $s_{v,t} = \phi(x_{v,t})$. These scores rank views by informativeness. To obtain a compact view-level importance measure, token-level scores are averaged across all tokens: $s_v = \frac{1}{T}\sum_{t=1}^{T} s_{v,t}$ where \(T\) denotes the number of tokens per view. The resulting scores $\{s_v\}_{v=1}^{V}$ rank view informativeness, and the Top-$K$ are selected: $\mathcal{V}_{\text{Top-}K} = \text{Top-}K(\{ s_v \}_{v=1}^{V})$. This encourages the network to focus on discriminative views (\cref{fig:teaser:a}, dominant views capture ~83\% of cumulative softmax attention weights during fusion), while suppressing redundant ones.

During training, we apply Gumbel-Softmax differentiable weighting~\cite{gumbelsoftmax} to enforce discriminative learning. At inference, the model switches to Soft Fusion to retain subtle complementary cues while filtering noise, effectively avoiding the information loss caused by hard pruning. Finally, features from the selected Top-$K$ views are aggregated into a unified exocentric representation:

\begin{equation}
x_{\text{exo}} = \text{Exos-Fusion}\Big(\{x_{v,t}\}_{v \in \mathcal{V}{\text{Top-}K}, t=1,\dots,T}\Big),
\end{equation}
where \(\text{Exos-Fusion}(\cdot)\) performs a weighted combination of the retained token features \({x_{v,t}}\) using their corresponding normalized importance scores \({s_{v,t}}\). The selection process is weakly supervised—scores are learned purely from task loss without any explicit view-level labels.
This operation is applied before the main transformer layers to prune redundant and uninformative views and highlight the most informative spatio-temporal cues. The resulting fused feature \(x_{\text{exo}}\) is subsequently passed into the exocentric decoder to produce token-level embeddings \(\{ x^{\text{exo}}_{t} \}_{t=1}^{T}\).

\paragraph{Egocentric Branch} 
The egocentric stream processes a single view and generates token embeddings \(\{x^{\text{ego}}_{t}\}_{t=1}^{T}\) via an egocentric decoder. These are fused with the exocentric representations 
\(\{x^{\text{exo}}_{t}\}_{t=1}^{T}\) through an EgoExo-Fusion framework, which reuses the Score Predictor to assign importance scores and adaptively weight ego–exo contributions at the token level. The unified embedding \(x_{\text{final}}\) is then fed into the Proficiency Estimator for classification.

\begin{equation}
x_{\text{final}} = \text{EgoExo-Fusion}\Big(\{x^{\text{ego}}_{t}\}_{t=1}^{T}, \{x^{\text{exo}}_{t}\}_{t=1}^{T}\Big).
\end{equation}
Here, $t$ denotes the token index, and $T$ is the number of tokens in the sequence.
\subsection{VIB-GB}\label{sec:3}

Even after pruning redundant views, fusion can overfit because the ego stream typically converges faster and dominates gradients. To address this, we regularise fusion through a two-part mechanism:
\begin{itemize}
\item \textbf{Gradient Blending regularisation (GB):} dynamically balances learning between ego and exo branches using the \emph{Overfitting-to-Generalisation Ratio (OGR)~\cite{wang2020makestrainingmultimodalclassification}}.
\item \textbf{Variational Information Bottleneck (VIB):} compress each branch’s latent space by limiting mutual information with the input, filtering out redundancy and keeping only task-relevant features.
\end{itemize}
\noindent\textbf{(a) OGR-guided Gradient Balancing.}
We use the \emph{Overfitting-to-Generalisation Ratio (OGR)}~\cite{wang2020makestrainingmultimodalclassification} to dynamically reweight gradients between ego and exo streams, down-scaling branches that overfit faster (see supp. for details). While the original OGR measures the ratio between these two factors, we adopt a simplified formulation that focuses solely on the overfitting component for training stability, as validation losses across views are often correlated.  
To compute OGR for each branch \(m\in\{\text{ego},\text{exo}\}\), we first quantify its degree of overfitting through the \emph{overfitting increment}:  
\begin{equation}
\Delta_O^{(m)}(e)=\mathcal{L}_{\text{train}}^{(m)}(e)-\mathcal{L}_{\text{val}}^{(m)}(e)
\end{equation}
at epoch \(e\), where \(\mathcal{L}_{\text{train}}\) and \(\mathcal{L}_{\text{val}}\) denote training and validation losses.
A large positive \(\Delta_O^{(m)}\) indicates that branch \(m\) is overfitting faster than it generalises. We therefore assign each branch an adaptive weight
\begin{equation}
w_m(e)=\frac{1}{\Delta_O^{(m)}(e)^p+\epsilon}, \qquad
w_m \leftarrow \frac{w_m}{\sum_j w_j},
\end{equation}
where \(p\in[0.5,1.0]\) controls sensitivity and \(\epsilon\) prevents division by zero. These weights reduce gradients for branches that overfit, equalising training dynamics.
\begin{equation}\label{eq:GB}
\mathcal{L}_{\text{GB}}=\sum_{m} w_m \mathcal{L}_{\text{train}}^{(m)}.
\end{equation}
During training, \(w_m\) and \(\Delta_O^{(m)}\) are updated online each epoch.
Intuitively, views that overfit more receive smaller gradients, keeping the ego and exo representations in sync and lowering the global OGR (see~\cref{fig:ogrwithvib} and~\cref{fig:ogrwovib}).

\noindent\textbf{(b) Variational Information Bottleneck.}
Even with balanced gradients, we found that the fused latent representation may still contain redundant signals.
To regularise it theoretically, we adopt the \emph{Variational Information Bottleneck (VIB)}~\cite{alemi2019deepvariationalinformationbottleneck} framework,
which constrains the mutual information between the input $X$ and the latent code $Z$ while maximising the information between $Z$ and the task label $Y$.
Formally, the information bottleneck objective can be expressed as:
\begin{equation}\label{eq:VIB_obj}
\max_{q_{\phi}(z|x)} \; \mathcal{I}(Z;Y) - \beta \mathcal{I}(Z;X),
\end{equation}
where $\mathcal{I}(\cdot;\cdot)$ denotes mutual information and $\beta$ controls the trade-off between sufficiency and compression.
In practice, we realise this objective using a lightweight bottleneck in each branch,
parameterised by mean and log-variance, and optimised through the reparameterisation trick.
The corresponding variational loss encourages each posterior $q(z_m|x_m)$ to align with a unit Gaussian prior $p(z)$:
\begin{equation}\label{eq:VIB}
\mathcal{L}_{\text{VIB}}=\sum_{m}
D_{\mathrm{KL}}\!\left(q(z_m|x_m)\,\|\,p(z)\right).
\end{equation}
This regularisation removes view-specific noise while retaining discriminative information,
yielding compact and generalisable cross-view representations. 

\begin{figure}
    \centering
    \includegraphics[width=0.60\linewidth]{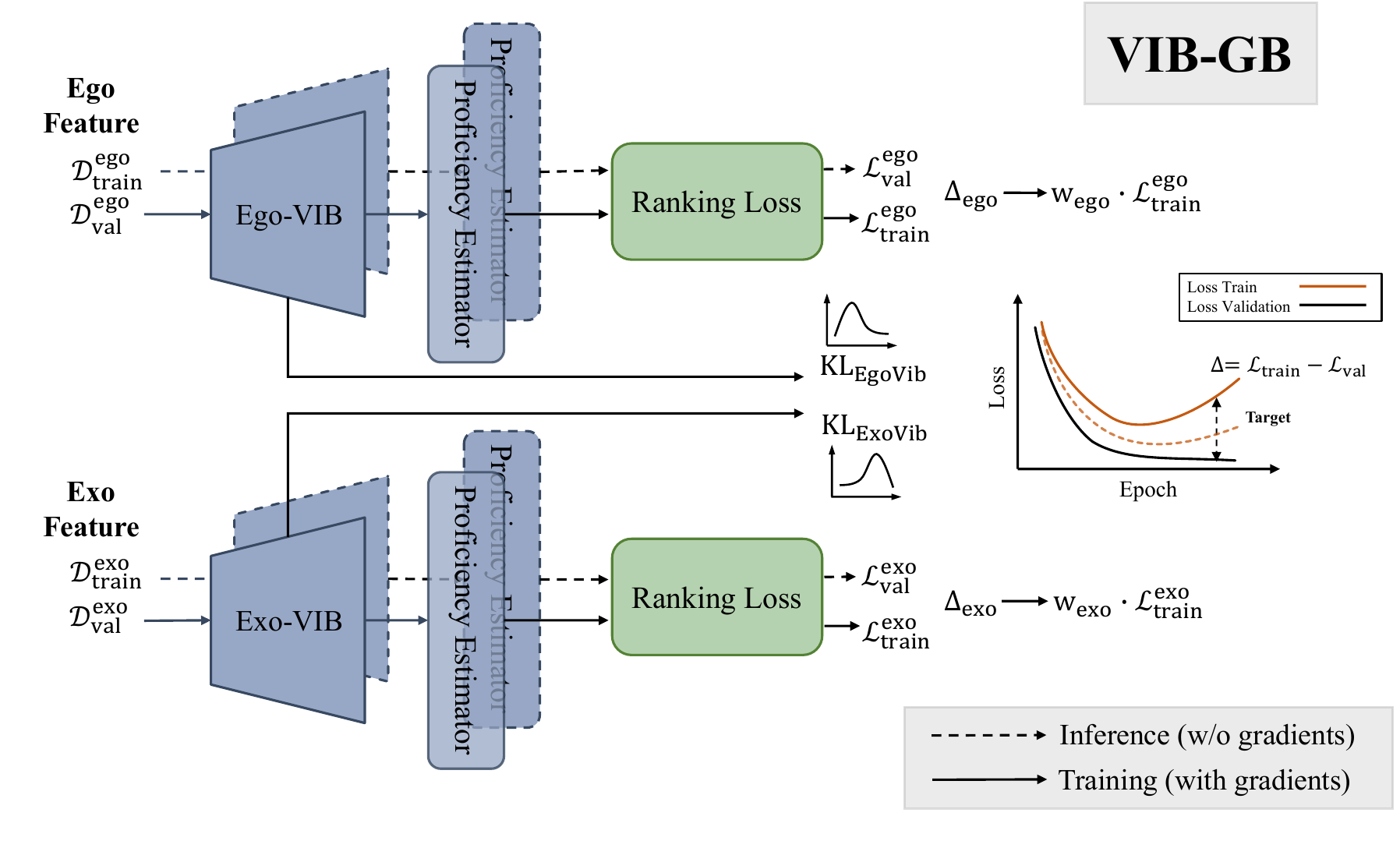}
    \caption{Overview of the proposed VIB-GB regulariser. Ego and exo features are regularised through separate VIB branches, and Gradient Blending Regularisation (GB) updates gradient weights online using both training and validation losses to balance learning and suppress overfitting. The KL term constrains the latent representation, while ${\Delta}_{ego/exo}$ guides proficiency learning.}
    \label{fig:VIBGB}
    \vspace{-20pt}
\end{figure}

Together, as illustrated in~\cref{fig:VIBGB}, the VIB-GB module operates with parallel \emph{ego} and \emph{exo} branches, each processing both training and validation inputs 
$\mathcal{D}_{\text{train}}$ and $\mathcal{D}_{\text{val}}$. 
Each branch passes its features through a VIB layer to obtain the KL term and computes the overfitting increment $\Delta_{ego/exo}$,
which together guide the OGR-based gradient blending for balanced optimisation across views. Unlike prior multimodal IB methods, VIB-GB couples information compression with dynamic OGR-based weighting, linking representational and training-dynamics regularisation.

\subsection{Optimisation}\label{sec:4}

Our model is optimised with a multi-objective loss combining three components. The ordinal loss provides ranking-based supervision for skill estimation. Gradient blending (GB) regularisation balances contributions from ego and exo modalities, while the variational information bottleneck (VIB) loss regularises the latent space to promote generalisable representations and mitigate overfitting.

We adopt CORN~\cite{shi2021deep} for ordinal classification, decomposing proficiency prediction into rank-consistent binary subtasks. CORN models action quality levels as ordered categories by decomposing the ordinal regression problem into a sequence of binary classification subtasks, thereby explicitly enforcing the inherent rank structure.
\begin{equation}
\mathcal{L}_{\text{CORN}} = - \tfrac{1}{N} \sum_{i=1}^{N} \sum_{k=1}^{K-1} \text{BCE}\Big(y_{i,k},\, \hat p_{i,k}\Big),
\end{equation}
where $\text{BCE}(\cdot)$ denotes the binary cross-entropy loss, $y_{i,k} \in \{0,1\}$ is the binary label indicating whether sample $i$ belongs to a level higher than $k$, $\hat p_{i,k}$ is the cumulative probability obtained via the chain rule, and $K$ is the total number of ordinal levels. Together with~\cref{eq:GB} and~\cref{eq:VIB}, these objectives ensure that proficiency estimation is both rank-aware (via CORN) and robust to redundant or overfit view signals (via VIB-GB).
\begin{equation}
\mathcal{L}_{\text{total}} =
\lambda_{\text{CORN}} \mathcal{L}_{\text{CORN}} +
\lambda_{\text{GB}} \mathcal{L}_{\text{GB}} +
\lambda_{\text{VIB}} \mathcal{L}_{\text{VIB}}.
\end{equation}
where $\lambda_{\text{GB}}$ and $\lambda_{\text{VIB}}$ are weighting coefficients applied during training. 
A moderate $\lambda_{\text{GB}}$ stabilises learning dynamics across ego and exo branches, 
while an appropriate $\lambda_{\text{VIB}}$ suppresses redundant view-specific signals and preserves discriminative cues.

In summary, at the \textbf{data level}, AdaMVS tackles redundancy through adaptive view selection, 
while at the \textbf{feature level}, VIB-GB mitigates overfitting through fusion regularisation, 
yielding compact and generalisable proficiency representations.

\section{Experiments}

\begin{table*}[htbp!]
\centering
\caption{Proficiency estimation accuracy (\%) comparison on the \textbf{Ego-Exo4D} and \textbf{EgoExo-Fitness} datasets in three settings. 
\textbf{Bold} numbers denote the best performance, and \underline{underlined} numbers denote the second best.}
\renewcommand{\arraystretch}{0.9}
\small

\begin{tabular}{l c ccc | ccc}
\toprule
\textbf{Method} & \textbf{Pretrain} & \multicolumn{3}{c}{\textbf{Ego-Exo4D} \cite{grauman2024egoexo4dunderstandingskilledhuman}} & \multicolumn{3}{c}{\textbf{EgoExo-Fitness} \cite{li2024egoexofitnessegocentricexocentricfullbody}} \\
\cmidrule(lr){3-5} \cmidrule(lr){6-8}
& & \textbf{Ego} & \textbf{Exos} & \textbf{Ego+Exos} & \textbf{Ego} & \textbf{Exos} & \textbf{Ego+Exos} \\
\midrule
Random & - & 26.4 & 26.4 & 26.4 & 26.2 & 26.2 & 26.2 \\
Majority-class & - & 32.3 & 32.3 & 32.3 & 33.3 & 33.3 & 33.3 \\
TimeSformer \cite{grauman2024egoexo4dunderstandingskilledhuman}& - & 40.6 & 39.0 & 39.9 & 32.1 & 32.1 & 32.1 \\
TimeSformer & EgoVLPv2\cite{pramanick2023egovlpv2} & 46.7 & 37.0 & 37.1 & 39.3 & 35.7 & 41.7 \\
TimeSformer & EgoVLP\cite{kevin2022egovlp} & 44.7 & 40.5 & 39.4 & 41.7 & 36.9 & 39.3 \\
TimeSformer & K400\cite{kay2017kineticshumanactionvideo} & 47.2 & 37.8 & 40.3 & 40.5 & 40.5 & 39.3 \\
TimeSformer & HowTo100M\cite{miech2019howto100mlearningtextvideoembedding}& 45.1 & 39.8 & 43.7 & 34.5 & 38.1 & 38.1 \\
SkillFormer \cite{bianchi2025skillformerunifiedmultiviewvideo} & K600\cite{carreira2018short} & 45.9 & 46.3 & 47.5 & 36.9 & 35.7 & 38.7 \\
\midrule
\textbf{Ours} & K600\cite{carreira2018short} & 48.6 & \underline{47.5} & 50.6 & 40.5 & 42.9 & \textbf{48.8} \\
\textbf{Ours} & K400\cite{kay2017kineticshumanactionvideo} & \underline{50.8} & \textbf{49.2} & \textbf{53.0} & \textbf{46.4} & \textbf{47.6} & \underline{46.4}\\
\textbf{Ours} & HowTo100M\cite{miech2019howto100mlearningtextvideoembedding} & \textbf{52.1} & \underline{47.5} & \underline{52.8} & \underline{42.9} & \underline{45.2} & 45.2 \\
\bottomrule
\end{tabular}
\label{tab:sota}
\end{table*}
We evaluate our method on two benchmarks: Ego-Exo4D for demonstrator proficiency estimation and EgoExo-Fitness 
for interpretable action judgement. We compare our model with state-of-the-art approaches under consistent 
protocols, and perform both quantitative and qualitative ablations to assess the contribution of each component. We train one unified model per dataset, handling all tasks simultaneously.

\noindent\textbf{Ego-Exo4D}~\cite{grauman2024egoexo4dunderstandingskilledhuman} is a large-scale multiview dataset spanning 9 activity categories (e.g., bouldering, cooking, music). It contains 1,087 hours of footage captured from one egocentric and four exocentric views, with 2,987 proficiency scores annotated across four discrete levels.

\noindent\textbf{EgoExo-Fitness} \cite{li2024egoexofitnessegocentricexocentricfullbody} contains 1,276 cross-view videos (approximately 32 hours) segmented into 6,131 single actions, annotated with interpretable action judgments and quality scores on a five-level scale.

\paragraph{Implementation Details}
All experiments are conducted on NVIDIA RTX 3090 GPUs. 
We adopt Timesformer \cite{grauman2024egoexo4dunderstandingskilledhuman} and VST \cite{liu2021video,liu2021Swin} as the backbone, using Kinetics-400/600 pretrained weights and HowTo100M pretrained weights.
Each video is uniformly sampled to 384 frames and divided into 8 consecutive-frame clips, resulting in 48 queries per video.
We train the model with a batch size of 64 using the Adam optimiser, with an initial learning rate of $1 \times 10^{-4}$, decayed by a factor of 0.1 every 30 epochs. 
Training is performed for 100 epochs with a dropout rate of 0.5. The weight for the variational information bottleneck term, $\lambda_{\mathrm{VIB}}$, is set to 0.1 (which is equivalent to $\beta$ in~\cref{eq:VIB_obj}); the weights of Corn Loss $\lambda_{CORN}$ and GB loss $\lambda_{GB}$ are all set to 1,
and the Top-$K$ view selection achieves the best performance when $K=2$.

\subsection{Comparison with State-of-the-Art}
We compare our model with current state-of-the-art approaches on the two benchmark datasets using three different pretrained backbones for video feature extraction.
As shown in~\cref{tab:sota}, we evaluate our framework under three settings: \textit{ego-only}, \textit{exo-only}, and \textit{ego+exo}.

Specifically, AdaMVS achieves a new state-of-the-art result on Ego-Exo4D, yielding a 5.5\% absolute improvement in overall accuracy compared to SkillFormer. To ensure a fair comparison under identical feature extraction settings, we further evaluate our framework using the same pretrained backbones as the baselines. Across all backbones (K400, K600, and HowTo100M), AdaMVS consistently delivers significant gains of 12.7\%, 3.1\%, and 9.1\%, respectively. On EgoExo-Fitness, our method obtains a 7.1\% absolute improvement.

We observe distinct view-importance patterns across the two benchmarks. 
In Ego-Exo4D, egocentric motion cues (e.g., Cooking, Music, Bouldering) play a more dominant role, with \textit{ego-only} configurations generally yielding higher accuracy. 
In contrast, the importance of views in EgoExo-Fitness (e.g., Fitness) is more balanced, where both \textit{ego-only} and \textit{exo-only} provide critical and complementary information. Previous methods often observe that combining ego and exo views degrades performance due to redundant or conflicting information. In contrast, our adaptive view-selection mechanism effectively mitigates this issue by dynamically weighting the most informative views and pruning redundant ones, leading to consistent and generalisable improvements across datasets.


\begin{table}[htbp!]
\centering
\footnotesize
\caption{Performance (\%) and model complexity comparison on the \textbf{Ego-Exo4D} datasets with different multiview and multi-modal fusion methods. \textbf{Bold} indicates the best performance, and \underline{underlined} indicates the second best.}
\renewcommand{\arraystretch}{0.9}
\begin{tabular}{@{\hskip 1mm}lccc|@{\hskip 1.3mm}c@{\hskip 1.3mm}c}
\toprule
\textbf{Method} & \textbf{Ego} & \textbf{Exos} & \textbf{Ego+Exos} & \textbf{GFLOPs} & \textbf{Params (M)} \\
\midrule
Summation           & 47.2 & 44.4 & 45.2 & 1.11 & 23.42 \\
Concat (early)      & 47.2 & 45.5 & 47.2 & 16.55 & 344.55 \\
Concat (late)       & 47.2 & \underline{47.7} & 49.0 & 11.74 & 23.42 \\
CrossFusion         & 47.2 & 47.5 & 46.8 & 1.67 & 4.20 \\
MTV \cite{yan2022multiview} & 45.2 & 43.7 & 43.5 & 1.15 & 7.37 \\
MVCNN \cite{su2015multiviewconvolutionalneuralnetworks} & 48.3 & 45.2 & 46.6 & 1.51 & 9.97 \\
MVAD \cite{he2024learning} & 46.3 & 46.3 & 45.0 & 0.56 & 5.25 \\
TokenFusion \cite{wang2022multimodaltokenfusionvision} & 47.0 & 45.5 & 49.0 & 1.41 & 13.65 \\
SkillFormer \cite{bianchi2025skillformerunifiedmultiviewvideo}& 45.9 & 46.3 & 47.5 & 1.25 & 20.60 \\
\midrule
\textbf{AdaMVS-Large}        & \textbf{50.8} & \textbf{49.2} & \textbf{53.0} & 2.98 & 24.47 \\
\textbf{AdaMVS-Small}& \underline{48.3} & 47.4 & \underline{50.1} & \textbf{0.26} & \textbf{2.19} \\
\bottomrule
\end{tabular}
\label{tab:egoexo_compare}
\vspace{-20pt}
\end{table}
\subsection{Comparison with General Fusion and Overfitting Baselines}
To further validate the effectiveness and efficiency of our AdaMVS fusion technique, we compare it with standard feature fusion methods and representative multimodal and multiview approaches. Since existing methods have not been evaluated on Ego–Exo datasets, we re-implemented them under our settings for fair comparison. For efficiency analysis, we introduce two variants, \emph{AdaMVS-Large} and \emph{AdaMVS-Small}. The small version incorporates a lightweight projection layer before the transformer modules, reducing parameters and FLOPs without degrading representation quality. As shown in~\cref{tab:egoexo_compare}, AdaMVS outperforms all compared methods. At the same time, the small variant achieves similar accuracy with only 0.26~GFLOPs and 2.19~M parameters—an over 11× reduction in computation and model size, demonstrating excellent efficiency with minimal performance loss. Furthermore, the comparison with different overfitting mitigation baselines is shown in~\cref{tab:overfittingmitation}. While generic regularisers apply stochastic perturbations, VIB-GB specifically targets branch-wise overfitting by balancing heterogeneous Ego-Exo learning dynamics, yielding superior performance. 

\begin{table}[b]
\centering
\footnotesize
\caption{Quantitative comparison against standard overfitting mitigation baselines with varying regularisation configurations.}
\label{tab:overfittingmitation}
\begin{tabular}{lc|c}
\hline
\textbf{Method (K400)} & \textbf{EgoExo4D} & \textbf{EgoExo-Fitness} \\ \hline
Baseline &  44.1 & 44.0\\ 
+ Feature-Mixup  & 45.2 & 40.5\\
+ Weight Decay (1e-2) & 44.6 & 44.0\\
+ Weight Decay (1e-3) &  44.1 & 44.0\\
+ Modality Dropout ($p=0.3$)  & 46.8 & 38.1 \\
+ Modality Dropout ($p=0.5$)  & 47.2  & 41.7\\
+ Standard Dropout ($p=0.3$)  & 47.5 & 38.1\\
+ Standard Dropout ($p=0.5$)  & 48.8 & 40.5\\ \hline
\textbf{Ours (+VIB-GB)} & \textbf{53.0} & \textbf{46.4} \\ \hline
\end{tabular}
\end{table}

\subsection{Ablation Study}

\subsubsection{Effectiveness of Different Proposed Modules}
In~\cref{tab:ablation}, we present the ablation study of the proposed modules.
For both the \textit{ego-only} and \textit{exo-only} settings, our AdaMVS module and Corn loss notably improve inter-view performance, yielding overall gains of around 3.0\% across both modalities.
In the \textit{ego–exo} setting, starting from the baseline (45.5\%), incorporating the AdaMVS module increases accuracy to 48.0\% (+2.5\%), demonstrating the effectiveness of adaptive view selection in filtering redundant or noisy views. Adding the ordinal loss $\mathbf{\mathcal{L}_{\text{CORN}}}$ further raises accuracy to 48.4\% (+0.4\%) by enforcing the natural order of skill levels and providing a stronger supervisory signal. Applying Gradient Blending (GB) alone yields 48.6\%, alleviating overfitting but still affected by redundant gradients from high-dimensional Ego–Exo features. When combined with the Variational Information Bottleneck (VIB), performance improves markedly to 51.0\% (+2.4\%), as VIB regularises the latent space and filters non-essential information. This final configuration demonstrates that all modules work synergistically to enhance overall performance.

We also conducted a decoupled evaluation of the VIB-GB module to isolate its contribution. When applied directly to the naive baseline, VIB-GB yields a 48.7\% accuracy (+3.2\%), confirming its effectiveness in mitigating the feature-level overfitting and enlarged feature space issues identified in our introduction. However, this result remains 2.3\% lower than our full model (51.0\%). This gap explicitly demonstrates that AdaMVS and VIB-GB are highly complementary: while AdaMVS suppresses uninformative or redundant views at the data level to move beyond brittle static fusion, VIB-GB prevents the model from memorising view-specific noise at the feature level.

\begin{table}[tp]
    \centering
    \footnotesize 
    
    \begin{minipage}{0.51\linewidth}
        \centering
        \footnotesize
        \caption{Ablation results of the proposed modules under \textit{Ego Only}, \textit{Exo Only}, and \textit{Ego+Exo} settings. \cmark\ denotes an enabled component, and \xmark\ denotes a disabled component. \textbf{Bold} indicates the best performance. Results are \textbf{averaged over five runs}.}
        \label{tab:ablation}
        \resizebox{\linewidth}{!}{%
        \begin{tabular}{lccccc}
        \toprule
        \textbf{Setting} & \textbf{AdaMVS} & $\mathbf{\mathcal{L}_{\text{CORN}}}$ & \textbf{GB} & \textbf{VIB} & \textbf{Acc} (\%) \\
        \midrule
        \textbf{\textit{ego-only}} & \xmark & \xmark & - & - & 46.8 \\
        \textbf{Ours} & \cmark & \cmark & - & - & \textbf{49.4} \\
        \midrule
        \textbf{\textit{exo-only}} & \xmark & \xmark & - & - & 45.1 \\
         \textbf{Ours} & \cmark & \cmark  & - & - & \textbf{48.6} \\
        \midrule
        \textbf{\textit{ego-exo}} & \xmark & \xmark & \xmark & \xmark & 45.5 \\
         & \cmark & \xmark & \xmark & \xmark & 48.0 \\
         & \cmark & \cmark & \xmark & \xmark & 48.4 \\
         & \cmark & \cmark & \cmark & \xmark & 48.6 \\ 
         \textit{W/O AdaMVS}& \xmark & \cmark & \cmark & \cmark &  48.7\\ 
         \midrule
        \textbf{Ours} & \cmark & \cmark & \cmark & \cmark & \textbf{51.0} \\
        \bottomrule
        \end{tabular}
        }
    \end{minipage}
    \hfill
    \begin{minipage}{0.46\linewidth}
        \centering
        \small
        \caption{Redundancy index $R_v$ (lower is better) and AdaMVS view weights (higher is better) for each action. 
For both metrics, the top-2 most informative views per action are highlighted in \colorbox{green!20}{green}. Results are \textbf{averaged over 200 runs}.}
        \label{tab:MI}
        \resizebox{\linewidth}{!}{%
        \begin{tabular}{l c cccc}
        \toprule
        \textbf{Action} & \textbf{Metric} & \textbf{exo1} & \textbf{exo2} & \textbf{exo3} & \textbf{exo4} \\
        \midrule
        Piano 
        & $R_v$\textdownarrow
        & 0.548 & 0.547 & \cellcolor{green!20}0.452 & \cellcolor{green!20}0.500 \\
        & AdaMVS\,\textuparrow
        & 0.000 & 0.000 & \cellcolor{green!20}0.470 & \cellcolor{green!20}0.530 \\
        \midrule
        Bouldering
        & $R_v$\textdownarrow
        & \cellcolor{green!20}0.356 & 0.383 & 0.390 & \cellcolor{green!20}0.362 \\
        & AdaMVS\,\textuparrow
        & \cellcolor{green!20}0.605 & 0.116 & 0.105 & \cellcolor{green!20}0.174 \\
        \midrule
        Soccer
        & $R_v$\textdownarrow
        & \cellcolor{green!20}0.418 & 0.420 & 0.419 & \cellcolor{green!20}0.402 \\
        & AdaMVS\,\textuparrow
        & 0.207 & 0.187 & \cellcolor{green!20}0.343 & \cellcolor{green!20}0.264 \\
        \midrule
        Dance
        & $R_v$\textdownarrow
        & \cellcolor{green!20}0.295 & 0.300 & \cellcolor{green!20}0.289 & 0.296 \\
        & AdaMVS\,\textuparrow
        & \cellcolor{green!20}0.301 & 0.189 & 0.209 & \cellcolor{green!20}0.301 \\
        \midrule
        Basketball
        & $R_v$\textdownarrow
        & \cellcolor{green!20}0.664 & \cellcolor{green!20}0.659 & 0.681 & 0.680 \\
        & AdaMVS\,\textuparrow
        & \cellcolor{green!20}0.323 & 0.136 & 0.193 & \cellcolor{green!20}0.348 \\
        \bottomrule
        \end{tabular}
        }
    \end{minipage}
\end{table}

\subsubsection{Effectiveness of AdaMVS and View Selection}
\begin{figure}[t]
    \centering
    \includegraphics[width=0.99\linewidth]{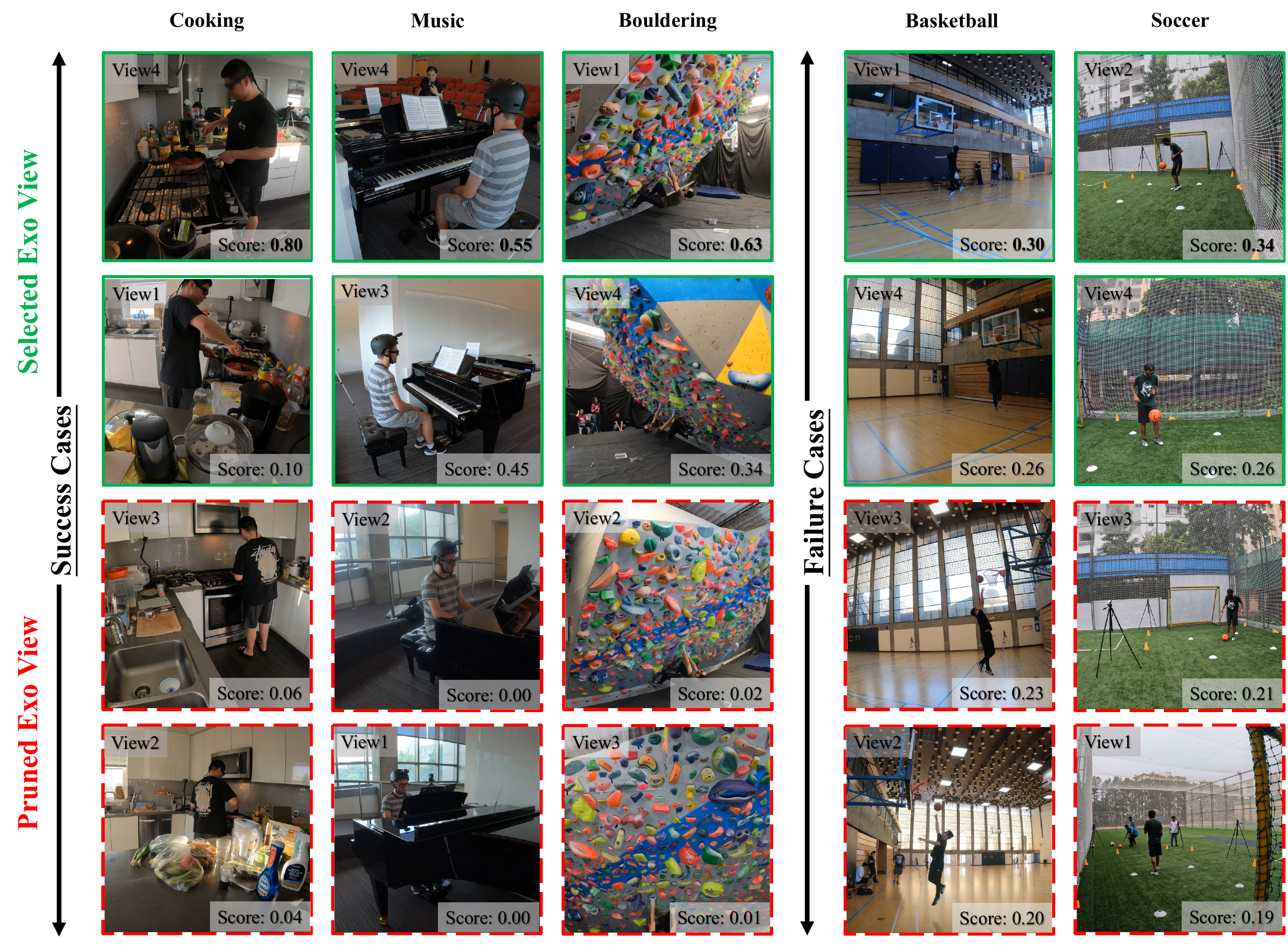}
    \caption{Visualisation of the view selection results from the AdaMVS module.
\textbf{Top}: success cases where the model selects the top-\emph{K} most informative exocentric views ($K=2$) to avoid multiview redundancy — green boxes denote selected views and red dotted boxes denote pruned ones.
\textbf{Bottom}: failure cases where all views receive similar scores, leading to suboptimal selection.}
\label{fig:viewselecvis}
\vspace{-15pt}
\end{figure}

We observe that incorporating more views does not necessarily improve performance, as some views may contain occlusions, motion blur, or irrelevant information.
As shown in the quantitative results in~\cref {tab:view_selection}, conventional fusion strategies often degrade as the number of views increases. To overcome this, our AdaMVS formulates view selection as a weakly supervised process that adaptively identifies the most informative views without explicit supervision. Qualitative results in~\cref{fig:viewselecvis} further show that predicted scores align with visual quality: low-scoring views often suffer from occlusions or blur, while high-scoring ones provide clearer and more discriminative cues.
In rare cases with highly overlapping content, the model assigns similar scores to all views, making it difficult to distinguish the most informative views.
\vspace{-10pt}
\paragraph{Data Redundancy Evaluation}
To analyse the correspondence between the redundancy patterns in the data and learned by AdaMVS, we measure each view’s task relevance and mutual similarity. Mutual Information (MI)~\cite{cover1999elements} estimates how much each view contributes to predicting proficiency, while the Centred Kernel 
Alignment (CKA)~\cite{kornblith2019similarity} captures the representation similarity between views. Based on these two factors, we define the redundancy index \( R_v = \frac{\mathrm{CKA\_avg}(v)}{\mathrm{MI}(v) + \epsilon} \), where a lower $R_v$ indicates a more informative and less redundant view. We further compare the learned AdaMVS weights with this redundancy index and observe a clear correspondence (\cref{tab:MI}): 
views with a lower redundancy index $R_v$ consistently receive higher importance weights. This confirms that 
AdaMVS effectively prioritises informative viewpoints while suppressing redundant ones, a trend also reflected in our visualisations (\cref{fig:viewselecvis}).

\paragraph{Comparison of K-Selection} 
We further compare the performance of AdaMVS under different values of $K$ for exocentric view sampling. Our AdaMVS adaptively selects the Top-$K$ informative exocentric views and filters out redundant ones before feature extraction. As shown in~\cref{tab:k_selection}, the model achieves the highest accuracy when $K=2$, validating the effectiveness of the adaptive view selection strategy.
This finding also aligns with the view distribution analysis in~\cref{fig:teaser:a}, where two dominant exocentric views account for approximately 82.8\% of all videos, and additional views mainly introduce redundant or noisy information.

\begin{table}[t]
    \centering
    \footnotesize
    
    \begin{minipage}{0.55\linewidth}
        \centering
        \caption{Comparison of different fusion methods on the EgoExo-4D dataset in terms of performance and efficiency under varying view configurations. 
    The $\pm$ values indicate variation across different view permutations, and \textbf{bold} numbers denote the best results.}
        \label{tab:view_selection}
        \resizebox{\linewidth}{!}{%
        \begin{tabular}{lccccc}
        \toprule
        \textbf{Fusion Method} & \textbf{1 Exo} & \textbf{2 Exo} & \textbf{3 Exo} & \textbf{4 Exo} & \textbf{Ego+Exos} \\
        \midrule
        Summation & $45.0$\textcolor{gray}{\scriptsize{$\pm$1.3}} & $45.5$\textcolor{gray}{\scriptsize{$\pm$2.0}} & $44.8$\textcolor{gray}{\scriptsize{$\pm$1.2}} & $44.4$ & $45.2$ \\
        Concat (early) & $45.0$\textcolor{gray}{\scriptsize{$\pm$1.3}} & $\mathbf{46.6}$\textcolor{gray}{\scriptsize{$\pm$0.9}} & $\mathbf{47.7}$\textcolor{gray}{\scriptsize{$\pm$1.5}} & $45.5$ & $47.2$ \\
        Concat (late) & $45.0$\textcolor{gray}{\scriptsize{$\pm$1.3}} & $45.7$\textcolor{gray}{\scriptsize{$\pm$1.6}} & $46.8$\textcolor{gray}{\scriptsize{$\pm$0.6}} & $47.7$ & $49.0$ \\
        Crossfusion & $45.0$\textcolor{gray}{\scriptsize{$\pm$1.3}} & $45.8$\textcolor{gray}{\scriptsize{$\pm$1.1}} & $47.1$\textcolor{gray}{\scriptsize{$\pm$0.8}} & $47.5$ & $46.8$ \\ \midrule
        \textbf{Ours (K600)} & $44.8$\textcolor{gray}{\scriptsize{$\pm$1.7}} & $46.1$\textcolor{gray}{\scriptsize{$\pm$2.0}} & $46.6$\textcolor{gray}{\scriptsize{$\pm$1.3}} & $\mathbf{49.0}$ & $\mathbf{50.6}$ \\
        \bottomrule
        \end{tabular}
        }
    \end{minipage}
    \hfill
    \begin{minipage}{0.44\linewidth}
        \centering
        \small
        \caption{Comparison of different $K$ selections for exocentric view sampling in AdaMVS on EgoExo-4D dataset. The best results for each dataset are highlighted in \textbf{bold}.}
        \label{tab:k_selection}
        \resizebox{\linewidth}{!}{%
        \begin{tabular}{lcccc}
        \toprule
        \textbf{Pretrained} & \textbf{$K=1$} & \textbf{$K=2$} & \textbf{$K=3$} & \textbf{$K=4$} \\
        \midrule
        K400        & $51.22$ & $\mathbf{53.00}$ & $51.00$ & $49.89$ \\
        K600        & $47.89$ & $\mathbf{50.55}$ & $48.56$ & $48.78$ \\
        HowTo100M   & $48.78$ & $\mathbf{52.77}$ & $51.22$ & $52.11$ \\
        \bottomrule
        \end{tabular}
        }
    \end{minipage}
    \vspace{-15pt}
\end{table}

\begin{figure}[b]
    \centering
    \captionsetup[subfigure]{font=tiny}
    \begin{subfigure}[t]{0.25\linewidth}
        \centering
        \includegraphics[width=\linewidth]{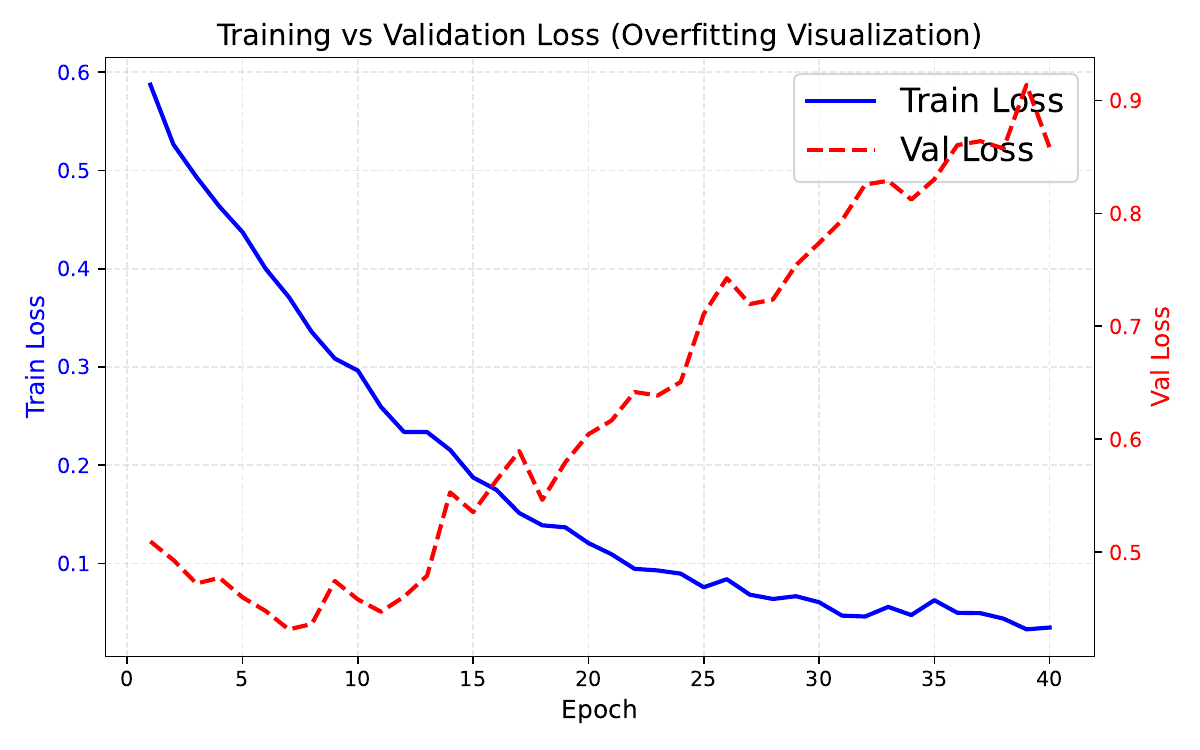}
        \caption{Train-Val w/o VIB-GB}
        \label{fig:trainwovib}
    \end{subfigure}
    \begin{subfigure}[t]{0.25\linewidth}
        \centering
        \includegraphics[width=\linewidth]{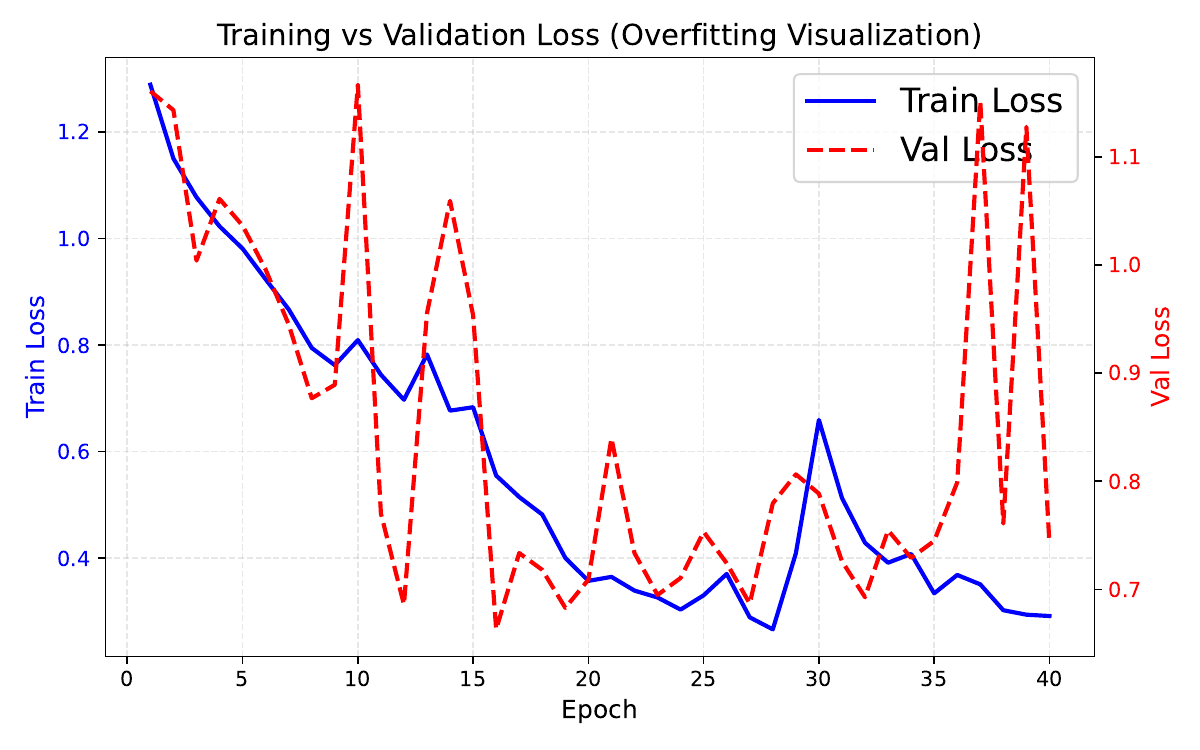}
        \caption{Train-Val with VIB-GB}
        \label{fig:trainwithvib}
    \end{subfigure}
    \begin{subfigure}[t]{0.235\linewidth}
        \centering
        \includegraphics[width=\linewidth]{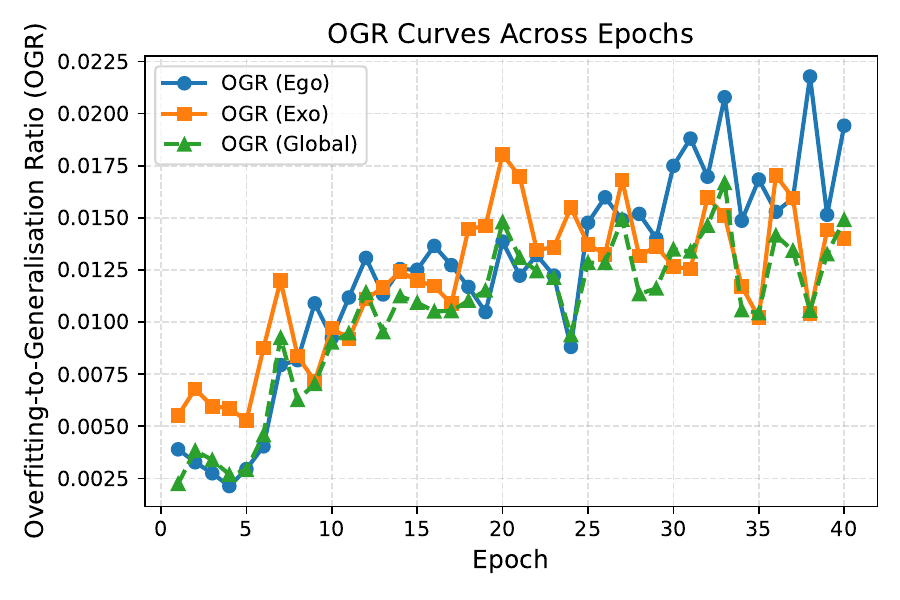}
        \caption{OGR w/o VIB-GB}
        \label{fig:ogrwovib}
    \end{subfigure}
    \begin{subfigure}[t]{0.235\linewidth}
        \centering
        \includegraphics[width=\linewidth]{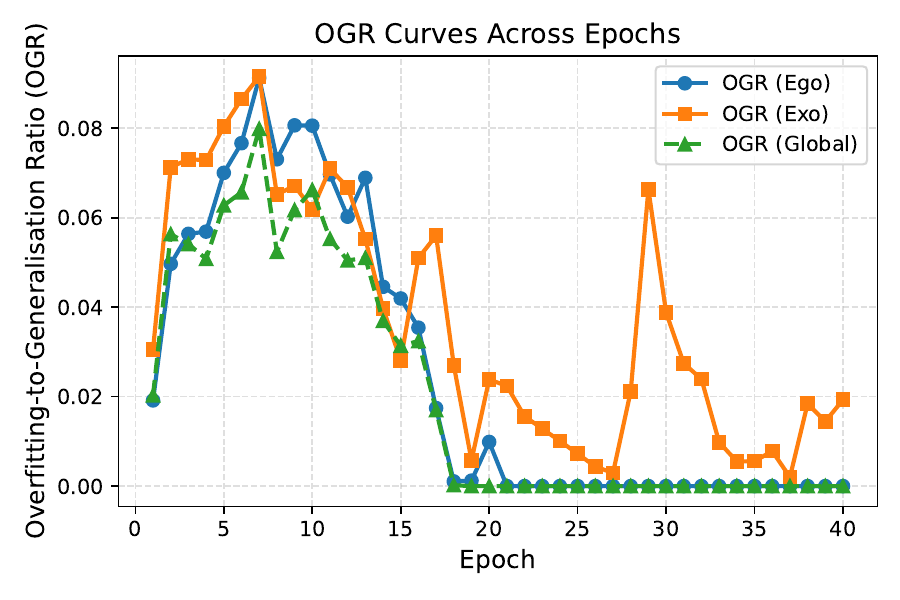}
        \caption{OGR with VIB-GB}
        \label{fig:ogrwithvib}
    \end{subfigure}
    \caption{\textbf{Analysis of overfitting and generalisation on the EgoExo-4D dataset.} 
\protect\subref{fig:trainwovib}--\protect\subref{fig:trainwithvib}: Training and validation curves of baseline without and with the proposed VIB-GB module, respectively. 
\protect\subref{fig:ogrwovib}--\protect\subref{fig:ogrwithvib}: Corresponding Overfitting-to-Generalisation Ratio (OGR) curves. 
Without VIB-GB, the model exhibits strong overfitting (\protect\subref{fig:trainwovib}--\protect\subref{fig:ogrwovib}), while our VIB-GB effectively regularises training dynamics and improves generalisation (\protect\subref{fig:trainwithvib}--\protect\subref{fig:ogrwithvib}).
}
    \label{fig:kinetics_overfitting}
\end{figure}

\subsubsection{OGR and VIB-GB Analysis}\label{sec:VIBGBexp}
To evaluate the model’s robustness against overfitting, we compare two configurations: the baseline model (with AdaMVS) and the model incorporating the proposed VIB-GB module, as shown in~\cref{fig:kinetics_overfitting}.
The top two subfigures (\cref{fig:trainwovib,fig:trainwithvib}) present the training and validation loss. The baseline starts to overfit around the 10th epoch, reaching 48.7\% accuracy, while the model with VIB-GB shows smoother training and a weaker overfitting trend, achieving 53.0\% accuracy, indicating that overfitting is effectively alleviated. The minor fluctuations in its validation reflect the epoch-wise updates of the VIB-GB component, which actively minimises overfitting during training.

The bottom two subfigures (\cref{fig:ogrwovib,fig:ogrwithvib}) show the OGR curves, which quantify the relative degree of overfitting across epochs. Without VIB-GB, OGR steadily increases, indicating progressive overfitting; in contrast, with VIB-GB, OGR rises slightly in the early stage (before the 10th epoch) and then decreases, stabilising after around 20 epochs, demonstrating its effectiveness in suppressing overfitting and improving generalisation. To further decouple the impacts of redundancy and overfitting, we conducted a controlled analysis using identical views and all-view configurations, with the detailed setup and results provided in the Supplementary Material.

\section{Conclusion}
In this work, we addressed Ego–Exo proficiency estimation and identified two key challenges in multiview fusion: redundant information and model overfitting. To tackle these issues, we proposed the Adaptive Multiview Selector (AdaMVS), which adaptively selects informative exocentric views at the data level, and the VIB-GB module, which integrates a variational information bottleneck with gradient blending at the feature level to alleviate overfitting. Empirical results demonstrate that these two modules are highly complementary, working together to enhance the overall robustness of the system.
Together, they form a general principle for adaptive fusion across heterogeneous (ego-exo) and homogeneous (exos) camera views. Moreover, the modular design of AdaMVS and VIB-GB allows the framework to extend to other multiview and multimodal tasks.
\paragraph{Limitations and future work}
Despite the challenges posed by the subjectivity of proficiency annotations and the limited scale of datasets, our framework demonstrates strong capability in mitigating these issues. Building on these promising results, we plan to extend the proposed approach to a broader range of multiview tasks in future work.

\bibliographystyle{splncs04}
\bibliography{main}
\end{document}


\title{Moving Beyond More Views: Redundancy-Aware Ego–Exo Fusion for Proficiency Estimation -- Supplementary File} 

\titlerunning{Redundancy-Aware Ego–Exo Fusion for Proficiency Estimation}

\author{Xu Dong\inst{1}\orcidlink{0009-0006-6842-8750} \and
Wanqing Li\inst{2}\orcidlink{0000-0002-4427-2687} \and
Anthony Adeyemi-Ejeye\inst{1}\orcidlink{0000-0002-8371-7829} \and
Andrew Gilbert\inst{1}\orcidlink{0000-0003-3898-0596}}

\authorrunning{X.Dong et al.}

\institute{University of Surrey, Guildford, United Kingdom\\ \and
University of Wollongong, Wollongong, Australia\\
\email{x.dong@surrey.ac.uk, a.gilbert@surrey.ac.uk}}

\maketitle

\appendix

\section{Additional Results}
\paragraph{Additional Visualizations}

\begin{figure}[htbp]
    \centering
    \includegraphics[width=0.95\linewidth]{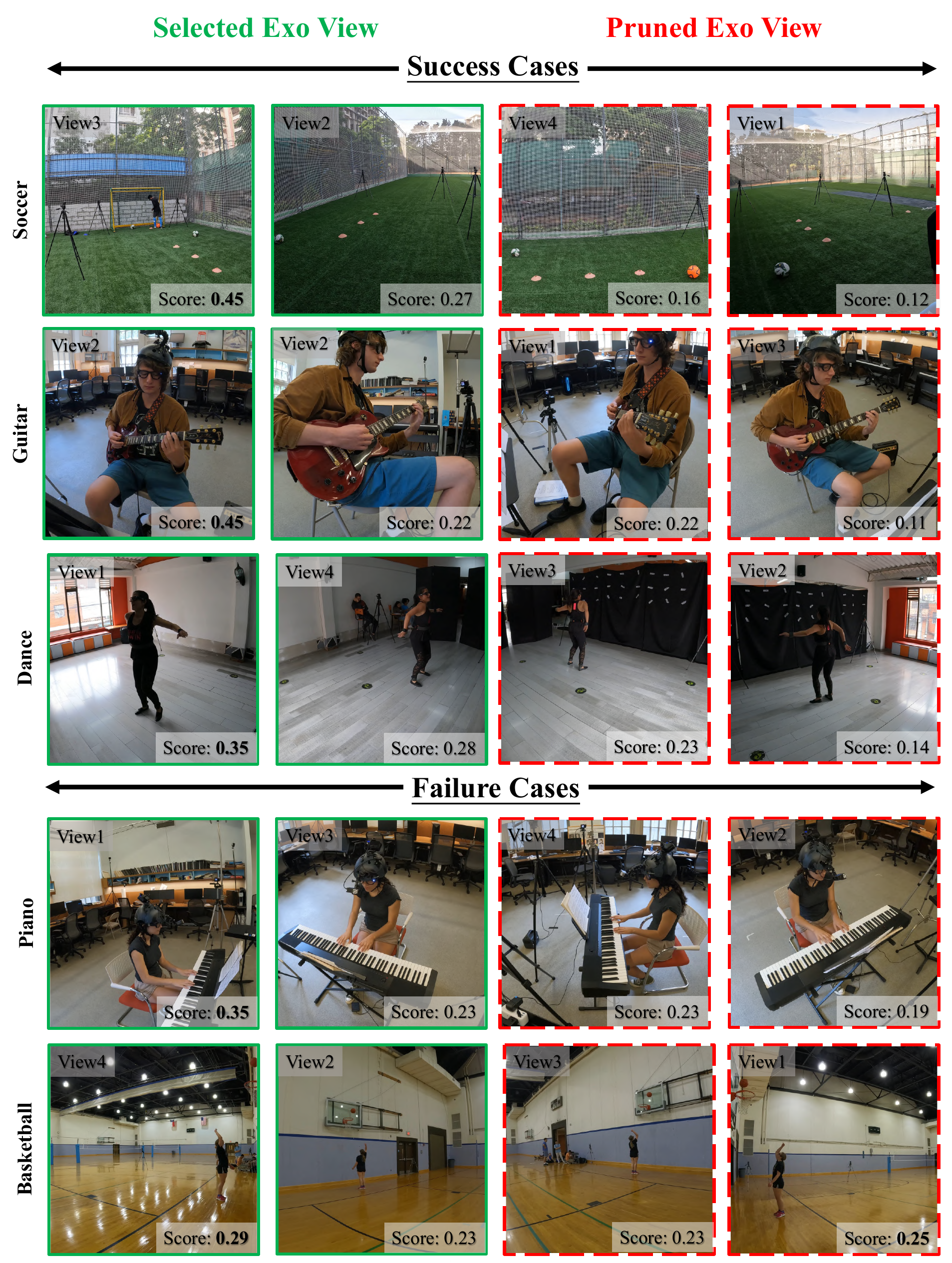}
    \caption{Visualisation of the view selection results from the AdaMVS module.
\textbf{Top}: success cases where the model selects the top-\emph{K} most informative exocentric views ($K=2$) to avoid multiview redundancy — green boxes denote selected views and red dotted boxes denote pruned ones.
\textbf{Bottom}: failure cases where all views receive similar scores, leading to suboptimal selection.}
    \label{fig:vis1}
\end{figure}

We provide additional qualitative visualisations of the view-importance scores predicted by AdaMVS (\cref{fig:vis1}).
In successful examples, AdaMVS assigns high weights to the most informative viewpoints while suppressing those that contribute little to proficiency reasoning — for instance, views suffering from occlusion, rear-facing viewpoints that capture limited motion cues, or perspectives dominated by background noise rather than the performer. This illustrates the model’s ability to prioritise discriminative spatio-temporal evidence and down-weight redundant or uninformative view tokens.

Conversely, in certain cases, the four exocentric views exhibit highly similar content or comparable visibility of the performer. Under such conditions, the model receives largely equivalent information from all views, resulting in relatively uniform predicted importance scores. This behaviour reflects the inherent ambiguity in redundancy assessment when view diversity is weak or when all views provide similarly informative signals.

\paragraph{View-wise Fusion Analysis}
\cref{tab:fusion_comparison_all_views_v15} reports the action recognition accuracy (\%) for a representative action label under different exocentric and egocentric view configurations. Each fusion method is evaluated using increasing numbers of available exocentric views (from one to four cameras), as well as in the ego–exo fusion setting. For the single-view configuration, results from individual cameras (1–4) are reported, while multi-view results summarise all possible permutations and their averaged performance.

Overall, the results indicate that conventional fusion strategies do not necessarily benefit from additional views; in some cases, performance even declines due to redundant or noisy visual information and the resulting overfitting. In contrast, our AdaMVS framework maintains stable gains as more views are introduced by adaptively selecting the most informative and complementary viewpoints for the given action instance. 


\begin{table*}[htbp]
\centering
\caption{
Detailed action recognition accuracy (\%) across all view configurations. 
Results are reported for each fusion method under varying numbers of exocentric views (1–4) and ego–exo combinations. `Avg.` denotes the mean accuracy and `SD` represents the standard deviation across corresponding view configurations.
}

\label{tab:fusion_comparison_all_views_v15}
\resizebox{\textwidth}{!}{%
\begin{tabular}{l|cccccc|cccccccc|cccccc|c|cc}
\toprule
\multirow{3}{*}{\textbf{Fusion Method}} 
& \multicolumn{6}{c|}{\textbf{1 Exo View}} 
& \multicolumn{8}{c|}{\textbf{2 Exo Views (6 Permutations)}} 
& \multicolumn{6}{c|}{\textbf{3 Exo Views (4 Permutations)}} 
& \multicolumn{1}{c|}{\textbf{4 Exo Views }} 
& \multicolumn{2}{c}{\textbf{Ego + Exo}} \\ 
\cline{2-24}
& \textbf{1} & \textbf{2} & \textbf{3} & \textbf{4} & \textbf{Avg.} & \textbf{SD}
& \textbf{12} & \textbf{13} & \textbf{14} & \textbf{23} & \textbf{24} & \textbf{34} & \textbf{Avg.} & \textbf{SD}
& \textbf{123} & \textbf{234} & \textbf{134} & \textbf{124} & \textbf{Avg.} & \textbf{SD}
& \textbf{1234} & \textbf{Avg.} & \textbf{SD} \\
\midrule
\textbf{Crossfusion}
& 44.12 & 43.68 & 45.68 & 46.56 & 45.01 & 1.34
& 46.12 & 46.12 & 44.12 & 47.01 & 46.34 & 44.79 & 45.75 & 1.08
& 47.01 & 46.34 & 46.78 & 48.12 & 47.06 & 0.76 & 47.45
& 46.78 & 1.34 \\
\textbf{Early Cat}
& 44.12 & 43.68 & 45.68 & 46.56 & 45.01 & 1.34
& 46.12 & 47.23 & 46.12 & 45.90 & 45.90 & 48.12 & 46.57 & 0.91
& 49.67 & 47.89 & 47.01 & 46.12 & 47.67 & 1.52 & 45.45
& 45.23 & 1.34 \\
\textbf{Late Cat}
& 44.12 & 43.68 & 45.68 & 46.56 & 45.01 & 1.34
& 45.45 & 46.12 & 46.34 & 43.46 & 48.12 & 44.57 & 45.68 & 1.60
& 46.56 & 46.12 & 47.01 & 47.45 & 46.79 & 0.57 & 47.67
& 49.00 & 1.34 \\
\textbf{Early Sum}
& 44.12 & 43.68 & 45.68 & 46.56 & 45.01 & 1.34
& 43.02 & 44.79 & 46.56 & 44.79 & 48.78 & 44.79 & 45.46 & 1.98
& 45.23 & 45.90 & 43.02 & 45.01 & 44.79 & 1.24 & 44.35
& 45.23 & 1.34 \\ \midrule
\textbf{Ours}
& 43.02 & 43.68 & 45.90 & 46.56 & 44.79 & 1.71
& 43.46 & 48.12 & 48.78 & 45.23 & 44.79 & 46.34 & 46.12 & 2.04
& 46.78 & 45.68 & 48.34 & 45.45 & 46.56 & 1.32 & 49.00
& \textbf{50.60} & \textbf{1.71} \\
\bottomrule
\end{tabular}%
}
\end{table*}

\paragraph{Redundancy/Overfitting Analysis}
To decouple and validate the claims of redundancy and overfitting, we conducted a controlled study (see ~\cref{tab:redundancy_study}): \textbf{Overfitting:} Adding identical views (Same Exo$\times$4) reduced performance, illustrating that increased model capacity without additional information triggers branch-wise overfitting. \textbf{Redundancy:} Using all Exos views also underperformed compared to our selective approach, confirming that redundant views introduce feature noise. Our AdaMVS achieves \textbf{49.2} by filtering redundancy and mitigating overfitting.
\begin{table}[h]
\centering
\footnotesize
\caption{Ablation and controlled analysis on view redundancy and model overfitting.}
\label{tab:redundancy_study}
\begin{tabular}{@{}l|ccc@{}}
\toprule
\textbf{Method} & \textbf{Exo} & \textbf{Same Exo$\times$4} & \textbf{All Exos} \\ \midrule
Summation       & 45.7 & 46.6 ($\uparrow$) & 44.4 ($\downarrow$) \\
Early Cat       & 45.7 &  43.0($\downarrow$) &  45.5($\downarrow$) \\
Late Cat        & 45.7 &  43.7($\downarrow$) &  47.7($\uparrow$) \\
Cross Fusion    & 45.7 &  43.7($\downarrow$) &  47.5($\uparrow$) \\ \midrule
\textbf{AdaMVS (Ours)} & \textbf{45.9} & \textbf{46.6} ($\uparrow$) & \textbf{49.2} ($\uparrow\uparrow$) \\ \bottomrule
\end{tabular}
\end{table}



\paragraph{Ego + Single-exo-view Comparison}
The per-view breakdown comparing Ego with each individual Exo view is shown in~\cref{tab:overfitting_baselines}, which shows that the dominant Exo view varies across different methods, proving that a single fixed view is insufficient. AdaMVS ensures task-agnostic robustness without manual tuning.

\begin{table}[h]
\centering
\footnotesize
\caption{Quantitative evaluation across different combinations of the egocentric view paired with individual exocentric views.}
\label{tab:overfitting_baselines}
\begin{tabular}{lcccc|c}
\hline
\textbf{Method} & \textbf{view1} & \textbf{+view2} & \textbf{+view3} &  \textbf{+view4} & \textbf{ego+exos} \\ \hline
Summation      & 43.9 & 47.2\textsubscript{\tiny 1st} & 44.4\textsubscript{\tiny 2nd} & 47.2\textsubscript{\tiny 1st} & 45.2 \\ 
Concat (early) & 45.7\textsubscript{\tiny 2nd} & 45.9\textsubscript{\tiny 1st} & 44.1 & 44.6 & 49.0 \\
Concat (late)  & 43.9\textsubscript{\tiny 2nd} & 44.6\textsubscript{\tiny 1st} & 43.7 & 43.7 & 47.2 \\
Crossfusion    & 46.1\textsubscript{\tiny 2nd} & 43.5 & 43.9 & 47.7\textsubscript{\tiny 1st} & 46.8 \\ \hline
\textbf{Ours}  & 46.1\textsubscript{\tiny 2nd} & 45.9 & 46.8\textsubscript{\tiny 1st} & 45.9 & 50.6 \\ \hline
\end{tabular}
\end{table}

\section{Metrics for Quantifying Multiview Redundancy}

Redundancy in multiview data arises from two complementary factors:
\begin{enumerate}
    \item \textbf{Low task informativeness} — the view contributes little to predicting proficiency.
    \item \textbf{High representational overlap} — the view encodes information already present in other views.
\end{enumerate} We quantify view redundancy using two complementary metrics. Mutual Information (MI) measures the task relevance of each view, while CKA captures representation-level similarity across views. Together, MI and CKA provide a principled measure of redundancy by evaluating both informativeness and inter-view overlap. 
\paragraph{Mutual Information (MI)}
Mutual Information (MI) measures how much a particular exocentric view $v$ reduces the uncertainty of the proficiency label $y$. Formally, MI quantifies the statistical dependency between two random variables:
\begin{equation}
\mathrm{MI}(v, y) = \iint p(v, y)\, \log \frac{p(v, y)}{p(v)\, p(y)}\, dv\, dy.
\end{equation}
In our setting, the feature embedding of each view serves as variable $v$, and the ordinal proficiency score corresponds to $y$.  
A higher MI indicates that the view provides stronger task-relevant cues for proficiency estimation, whereas a low MI suggests that the view is noisy, weak, or uninformative.  
Thus, MI captures \textbf{task relevance} of each view.

\subsection*{Centred Kernel Alignment (CKA)}
Centred Kernel Alignment (CKA) quantifies the similarity between two sets of feature representations.  
Given representations $X$ and $Z$ from two views, the linear CKA is:
\begin{equation}
\mathrm{CKA}(X, Z) =
\frac{\| Z^\top X \|_{F}^{2}}
     {\| X^\top X \|_{F}\, \| Z^\top Z \|_{F}}.
\end{equation}
CKA ranges from $0$ (dissimilar) to $1$ (identical up to rotation).  
A high CKA between views indicates overlapping or redundant information content.  
Due to its invariance to isotropic scaling and orthogonal transformations, CKA is a reliable tool for diagnosing redundancy among neural representations.

To characterise redundancy from both perspectives, we define a redundancy index:
\begin{equation}
R_v = \frac{\mathrm{CKA\_avg}(v)}{\mathrm{MI}(v) + \epsilon},
\end{equation}
where $\mathrm{CKA\_avg}(v)$ denotes the mean CKA similarity between view $v$ and the remaining views, and $\epsilon$ avoids division by zero.  
A lower $R_v$ indicates a more informative and less redundant view; a higher $R_v$ indicates that the view is duplicative or noisy.

Empirically, we observe a strong correspondence between the redundancy index $R_v$ and the learned AdaMVS importance weights:  
views with low redundancy (\emph{low} $R_v$) consistently receive higher weights. This demonstrates that AdaMVS naturally prioritises informative and complementary views, aligning with redundancy patterns present in the data.

\subsection{Overfitting-to-Generalisation Ratio (OGR)}

\begin{figure}[h]
    \centering
    \includegraphics[width=0.5\linewidth]{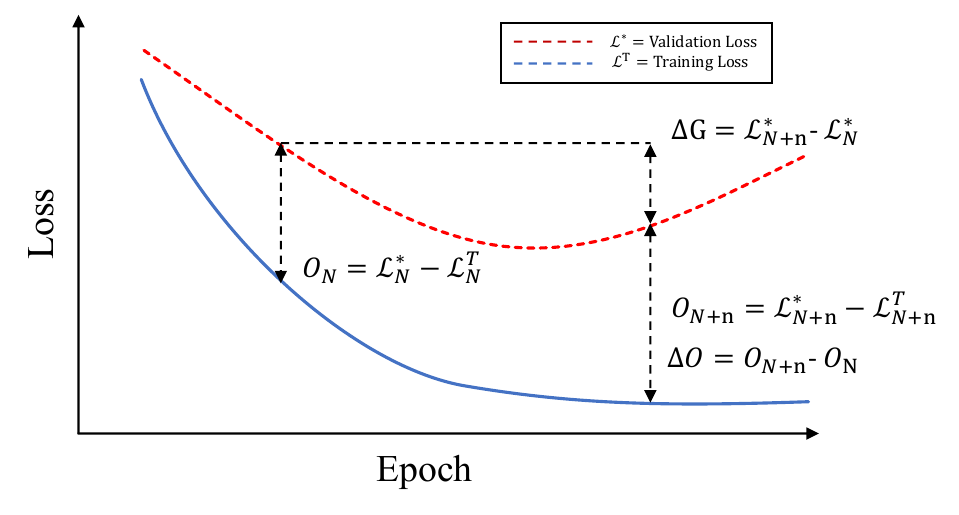}
    \caption{
    Illustration of the Overfitting-to-Generalisation Ratio (OGR) concepts during training.}
    \label{fig:ORG}
\end{figure}

To quantify how training dynamics relate to overfitting,
\cite{wang2020makestrainingmultimodalclassification} introduced the
\textit{Overfitting-to-Generalisation Ratio (OGR)}, which measures how much
overfitting increases relative to the improvement in generalisation (~\cref{fig:ORG}). Formally:
\begin{equation}
\mathrm{OGR}_{N,n} =
\frac{|\Delta \mathcal{O}_{N,n}|}{|\Delta \mathcal{G}_{N,n}|}
=
\frac{|\mathcal{O}_{N+n}-\mathcal{O}_{N}|}
     {|\mathcal{L}^{*}_{N+n}-\mathcal{L}^{*}_{N}|},
\label{eq:OGR}
\end{equation}
where:
\begin{itemize}[leftmargin=*]
    \item $\mathcal{L}^{T}$ and $\mathcal{L}^{*}$ denote training and validation loss respectively
    (validation serves as a proxy for the true generalisation error);
    \item $\mathcal{O}_{N} = \mathcal{L}^{*}_{N} - \mathcal{L}^{T}_{N}$ is the overfitting value at iteration $N$;
    \item $\Delta \mathcal{G}_{N,n} = \mathcal{L}^{*}_{N+n} - \mathcal{L}^{*}_{N}$ denotes the validation-loss decrease
    (generalisation gain);
    \item $\Delta \mathcal{O}_{N,n} = \mathcal{O}_{N+n} - \mathcal{O}_{N}$ measures the increment of overfitting.
\end{itemize}

A higher OGR indicates that optimisation is dominated by overfitting rather than genuine generalisation improvement.

\paragraph{Gradient Blending (GB)}
Based on this measure, \textit{Gradient Blending} \cite{wang2020makestrainingmultimodalclassification}, which adaptively fuses per-modality gradients to suppress overfitting.
Each modality gradient $\mathbf{g}_{k}$ is weighted according to its contribution
to generalisation relative to its overfitting variance:
\begin{equation}
\mathbf{g}_{\text{blend}}
=
\sum_{k=1}^{K} w_k\, \mathbf{g}_k,
\qquad
w_k \propto \frac{\langle \nabla \mathcal{L},\, \mathbf{g}_k\rangle}{\sigma_k^2},
\label{eq:GB}
\end{equation}
where $\sigma_k^2$ denotes the overfitting variance of modality $k$.
Gradients that generalise well (low OGR) receive larger weights, while gradients prone to overfitting are down-weighted.

Although the original OGR formulation jointly considers the overfitting increment $\Delta O$ and the generalisation gain $\Delta G$, we empirically found that the $\Delta G$ term provides limited discriminative value in our setting.
Specifically, across Ego--Exo view configurations, the validation losses of different views exhibit high inter-view correlation, leading to nearly identical $\Delta G$ trajectories for all branches. As a result, $\Delta G$ fails to reflect view-specific generalisation behaviour and instead introduces additional instability due to noise in validation-loss fluctuations.

In contrast, the overfitting increment $\Delta O$ shows strong and consistent separation between ego and exo branches—ego features typically converge faster and overfit earlier—making $\Delta O$ a more reliable indicator of modality-specific overfitting.
Consistent with these findings, we adopt a simplified OGR variant that focuses solely on $\Delta O$ for training stability, ensuring that gradient reweighting responds directly to the view that overfits more rapidly.

\section{Methodology}
\subsection{Transformer Decoder Structure}
\begin{figure}
    \centering
    \includegraphics[width=0.5\linewidth]{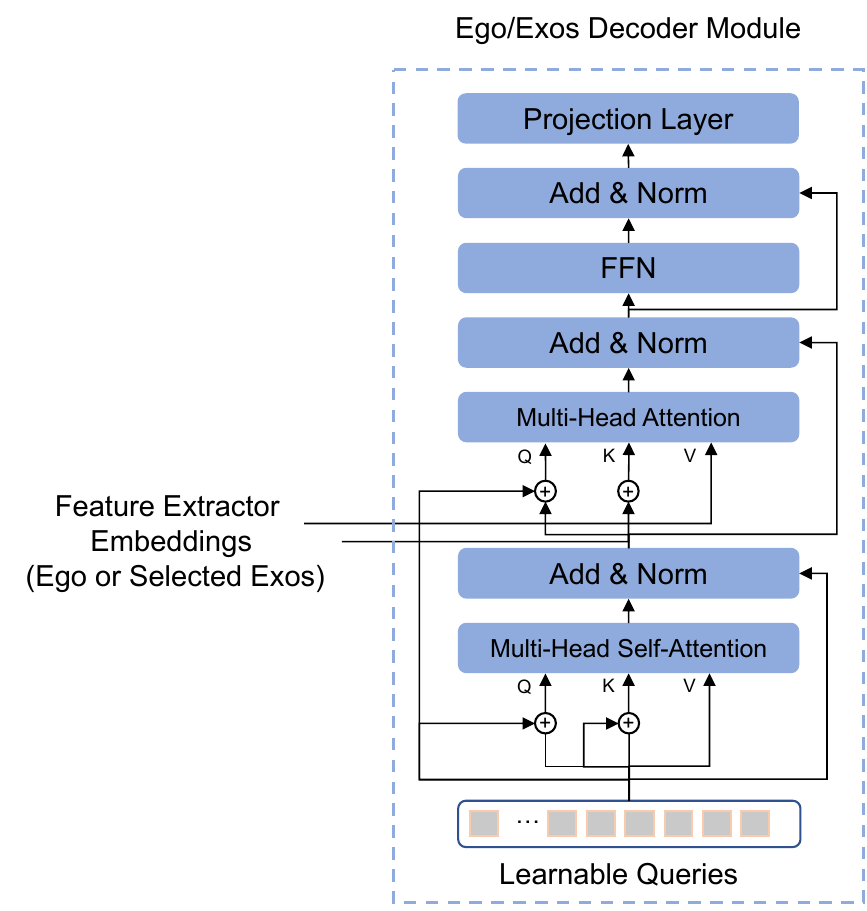}
    \caption{This module is a Transformer Decoder structure which refines Learnable Queries via Self-Attention before fusing them with Feature Extractor Embeddings via Multi-Head Attention.}
    \label{fig:detr}
\end{figure}

Our decoder follows a DETR-style \cite{detr} Transformer architecture that incorporates learnable queries as the core decoding tokens, as shown in \cref{fig:detr}.
Each decoder layer contains a multi-head self-attention module to model token dependencies among queries, followed by a feed-forward network (FFN) to refine their representations.
Residual connections and layer normalisation are applied after each sub-layer to ensure stable optimisation.
The learnable query embeddings interact with encoded feature sequences through self-attention, allowing the decoder to adaptively attend to relevant spatio-temporal information and generate structured, task-specific representations.

\subsection{AdaMVS Pseudocode}
To provide a clear overview of our model’s internal workflow, we present the pseudocode of the two core components in our framework — the Adaptive Multi-View Selection (AdaMVS) modules for both the exocentric and egocentric branches.
The AdaMVS design enables our system to dynamically evaluate and select informative views or tokens at different stages of the fusion process.
Specifically, the exocentric branch focuses on adaptively selecting the most discriminative external camera views to construct a compact, high-quality multi-view representation, while the egocentric branch further integrates egocentric and exocentric cues through token-level adaptive weighting for robust final embedding generation.
The detailed computational procedures for these two branches are summarised in Algorithms~\cref{alg:exo_branch} and~\cref{alg:ego_branch}.

\begin{algorithm}[h]
\caption{AdaMVS Exocentric Branch}
\label{alg:adamvs}
\KwIn{Tokens $\{x_{v,t}\}$ from $V$ views}
\KwOut{Fused exocentric tokens $\{x^{\text{exo}}_t\}$}

\textbf{1. Token scoring:}
$s_{v,t}=\phi(x_{v,t})$

\textbf{2. Gumbel–Softmax weighting (Train \& Test):}
$w_{t,b,v}=\mathrm{GumbelSoftmax}(s_{t,b,v};\tau,\texttt{hard=False})$

\textbf{3. Train (Top-$K$ masking):}
Compute per-view scores by averaging token weights:
$\bar{s}_{b,v}=\frac{1}{T}\sum_t w_{t,b,v}$;

Select the Top-$K$ most informative views:
$\mathcal{V}^{(b)}_{\text{top-}K}=\mathrm{TopK}(\bar{s}_{b,:})$;

Construct a binary mask for selected views:
$m_{t,b,v}=\mathbb{I}[\,v\in\mathcal{V}^{(b)}_{\text{top-}K}\,]$;

Apply the mask and normalise weights among Top-$K$:
$\hat{w}_{t,b,v}=\frac{m_{t,b,v}w_{t,b,v}}
     {\sum_{v'}m_{t,b,v'}w_{t,b,v'}}$;

Fuse the selected views using the renormalised weights:
$x^{\text{exo}}_{t,b}=\sum_v \hat{w}_{t,b,v}x_{v,t,b}$.

\textbf{4. Test (soft fusion, no Top-$K$):}
$\tilde{w}_{t,b,v}=\frac{w_{t,b,v}}{\sum_{v'} w_{t,b,v'}}$;  
$x^{\text{exo}}_{t,b}=\sum_v \tilde{w}_{t,b,v}x_{v,t,b}$.

\textbf{5. Exocentric decoding:}
$x^{\text{exo}}_{t,b}=\mathrm{ExoDecoder}(x^{\text{exo}}_{t,b})$.
\label{alg:exo_branch}
\end{algorithm}

\begin{algorithm}[h]

\caption{Ego–Exo Fusion with Adaptive Modality Weighting}
\label{alg:egoexo}
\KwIn{Ego tokens $x^{\text{ego}}_{t,b}$\; Exo tokens $x^{\text{exo}}_{t,b}$}
\KwOut{Fused tokens $\{x^{\text{final}}_{t,b}\}$}

\BlankLine
\textbf{1. Feature projection} \\
Project both modalities:
$x^{\text{ego}}_{t,b}=\mathrm{Proj}_{\text{ego}}(x^{\text{ego}}_{t,b}),\;
 x^{\text{exo}}_{t,b}=\mathrm{Proj}_{\text{exo}}(x^{\text{exo}}_{t,b})$.

\BlankLine
\textbf{2. Modality scoring} \\
Predict ego/exo importance:
$\alpha_{t,b}=\phi_{\text{ego}}(x^{\text{ego}}_{t,b}),\;
 \beta_{t,b}=\phi_{\text{exo}}(x^{\text{exo}}_{t,b})$, \\
forming scores $S_{t,b}=[\alpha_{t,b},\beta_{t,b}]$.

\BlankLine
\textbf{3. Adaptive weighting} \\
\textbf{Training:} average scores over tokens,
$\bar{S}_{b}=\tfrac{1}{T}\sum_t S_{t,b}$, \\
apply Gumbel–Softmax
$w_{b}=\mathrm{GumbelSoftmax}(\bar{S}_{b};\tau)$, \\
broadcast to all tokens and fuse:
$x^{\text{final}}_{t,b}=\sum_{m\in\{\text{ego},\text{exo}\}} w_{b,m} x^{m}_{t,b}$.

\textbf{Testing:} normalise scores,
$\tilde{w}_{t,b,m}=S_{t,b,m}/\sum_{m'}S_{t,b,m'}$, \\
soft fusion of both modalities:
$x^{\text{final}}_{t,b}=\sum_{m} \tilde{w}_{t,b,m} x^{m}_{t,b}$.

\BlankLine
\textbf{4. Output} Return $\{x^{\text{final}}_{t,b}\}$ for downstream decoding.
\label{alg:ego_branch}
\end{algorithm}

\section{Extended Related Work}

\paragraph{Egocentric–Exocentric Correlation.}
While the main paper focuses on adaptive cross-view fusion, prior work has explored the ego–exo relationship from broader perspectives, including view transfer, geometric alignment, and correspondence learning. 
Earlier studies~\cite{li2021egoexotransferringvisualrepresentations,lifting} aim to transfer motion patterns or embeddings between viewpoints through geometric constraints or contrastive objectives. 
Other methods seek to align latent representations across perspectives via adversarial or metric learning schemes, reducing domain gaps without explicit correspondence labels. 
Recent advances extend this to temporal alignment~\cite{xue2023learningfinegrainedviewinvariantrepresentations}, capturing frame-level correspondence to encourage view-consistent representations.
However, these works generally assume equal informativeness across views and do not account for redundancy or ambiguity within view streams—issues our adaptive selection module explicitly mitigates.  

\paragraph{Cross-view Skill Analysis.}
Beyond the EgoExo4D benchmark, several domains have adopted cross-view reasoning for skill and behaviour understanding. 
For instance, instructional video analysis and sports assessment often employ mixed first-person and third-person footage to improve robustness and interpretability. 
However, most frameworks~\cite{zhou2025uncertaintydrivenactionqualityassessment,doughty2019prosconsrankawaretemporal} rely on temporally aggregated features and static fusion, neglecting token-level variability in view quality.
In contrast, our token-level AdaMVS design dynamically weighs each view and temporal token, offering finer control over redundant or noisy visual cues during skill evaluation.

\paragraph{Cross-modal and Multi-view Fusion.}
Beyond ego–exo understanding, multi-view and multi-modal fusion have been studied across diverse domains such as 3D reconstruction~\cite{hamdi2021mvtnmultiviewtransformationnetwork,su2015multiviewconvolutionalneuralnetworks}, remote sensing, and multimodal perception.
Traditional fusion strategies often operate at fixed levels—early (feature concatenation), intermediate (attention-based fusion), or late (score-level averaging)—but struggle to adaptively model feature relevance over time.
Recent transformer-based methods~\cite{wang2022multimodaltokenfusionvision,jia2024geminifusionefficientpixelwisemultimodal} introduce token-wise interaction between modalities, but their complexity and data requirements limit scalability in ego–exo scenarios with asynchronous camera streams.
Our framework draws inspiration from these fine-grained paradigms but introduces view adaptivity and view-level regularisation to control redundancy and overfitting.

\paragraph{Generalisation and Overfitting in Multi-view Learning.}
A subtle yet critical challenge in multi-view systems is generalisation imbalance: different views tend to overfit or underfit at different rates~\cite{wang2020makestrainingmultimodalclassification,tejero2024complementary}.
While prior work quantified this phenomenon through the Overfitting-to-Generalisation Ratio (OGR), few attempted to explicitly optimise it. 
Our integration of the VIB-GB regulariser complements AdaMVS by suppressing view-specific overfitting, balancing the learning pace between ego and exo streams, and producing more stable cross-view representations.

\section{Limitations and Future Work}
Although AdaMVS learns view‐importance scores in a weakly supervised manner, the supervision signal ultimately derives from proficiency annotations, which are inherently subjective and often ambiguous. This noisiness may limit the precision of the learned importance scores, especially for actions where annotators disagree or where skill cues are subtle. 

Moreover, although the adaptive Top-K mechanism helps mitigate cases where multiple views contribute similarly, it cannot fully resolve them. For samples with simple motions or highly overlapping camera coverage, different views provide nearly equivalent information, causing the model to assign almost uniform weights. In such cases, AdaMVS cannot reliably distinguish a clearly dominant view, reflecting an inherent limitation of the data rather than the method.

In the future, we plan to explore adaptive K-selection strategies further, allowing the number of selected views to adjust dynamically based on action complexity and cross-view variability. We also aim to extend proficiency estimation to more advanced multimodal reasoning tasks, such as integrating Ego–Exo action understanding with language-based feedback generation, enabling more comprehensive and interpretable human performance analysis.

\bibliographystyle{splncs04}
\bibliography{main}